\documentclass[10pt,journal,compsoc]{IEEEtran}

\ifCLASSOPTIONcompsoc
  \usepackage[nocompress]{cite}
\else
  \usepackage{cite}
\fi

\usepackage{graphicx}
\usepackage{amsmath,amssymb} 
\usepackage{color}
\usepackage{wrapfig}
\usepackage{makecell}
\usepackage{svg}
\usepackage{rotating}
\usepackage{enumitem}
\usepackage{multirow}
\usepackage{microtype}
\usepackage{xparse}
\usepackage{booktabs,tabu}
\usepackage{floatrow}

\usepackage{hyperref}

\newcommand\degr[0]{^{\circ}}
\ExplSyntaxOn
\newcommand\latinabbrev[1]{
  \peek_meaning:NTF . {
    #1\@}%
  { \peek_catcode:NTF a {
      #1.\@ }%
    {#1.\@}}}
\ExplSyntaxOff

\newcommand{\boldomega}{\boldsymbol \omega} 
\newcommand{\boldmu}{\boldsymbol \mu} 
\newcommand{\bolddelta}{\boldsymbol \delta} 

\newcommand{\boldrho}{\boldsymbol \rho}
\newcommand{\Reals}{\mathbb{R}}

\newcommand{\ie}{\latinabbrev{i.e}}
\newcommand{\eg}{\latinabbrev{e.g}}

\graphicspath{{figures/}}

\begin{document}

\title{Single Day Outdoor Photometric Stereo}

\author{Yannick Hold-Geoffroy\textsuperscript{$\ddagger$},\: Paulo Gotardo\textsuperscript{$\dagger$},\: Jean-Fran\c{c}ois Lalonde\textsuperscript{*}
\IEEEcompsocitemizethanks{\IEEEcompsocthanksitem Adobe\textsuperscript{$\ddagger$}, Disney Research Studios\textsuperscript{$\dagger$}, Universit\'e Laval\textsuperscript{*}}
}



\IEEEtitleabstractindextext{%
\begin{abstract}
Photometric Stereo (PS) under outdoor illumination remains a challenging, ill-posed problem due to insufficient variability in illumination. Months-long capture sessions are typically used in this setup, with little success on shorter, single-day time intervals. In this paper, we investigate the solution of outdoor PS over a single day, under different weather conditions. First, we investigate the relationship between weather and surface reconstructability in order to understand when natural lighting allows existing PS algorithms to work. Our analysis reveals that partially cloudy days improve the conditioning of the outdoor PS problem while sunny days do not allow the unambiguous recovery of surface normals from photometric cues alone. We demonstrate that calibrated PS algorithms can thus be employed to reconstruct Lambertian surfaces accurately under partially cloudy days. Second, we solve the ambiguity arising in clear days by combining photometric cues with prior knowledge on material properties, local surface geometry and the natural variations in outdoor lighting through a CNN-based, weakly-calibrated PS technique. Given a sequence of outdoor images captured during a single sunny day, our method robustly estimates the scene surface normals with unprecedented quality for the considered scenario. Our approach does not require precise geolocation and significantly outperforms several state-of-the-art methods on images with real lighting, showing that our CNN can combine efficiently learned priors and photometric cues available during a single sunny day. 
\end{abstract}

}

\maketitle
\IEEEdisplaynontitleabstractindextext
\IEEEpeerreviewmaketitle


\IEEEraisesectionheading{\section{Introduction}\label{sec:introduction}}
\label{introduction}

\begin{figure*}
    \centering
    \setlength{\tabcolsep}{0pt} 
    \newcommand{\customwidth}{.079\linewidth}
    \begin{tabular}{@{}rcccccccccccc@{}}
                                                     &
    \begin{minipage}{\customwidth}\centering\scriptsize 11:00 \end{minipage} &
    \begin{minipage}{\customwidth}\centering\scriptsize 11:30 \end{minipage} &
    \begin{minipage}{\customwidth}\centering\scriptsize 12:00 \end{minipage} &
    \begin{minipage}{\customwidth}\centering\scriptsize 12:30 \end{minipage} &
    \begin{minipage}{\customwidth}\centering\scriptsize 13:00 \end{minipage} &
    \begin{minipage}{\customwidth}\centering\scriptsize 13:30 \end{minipage} &
    \begin{minipage}{\customwidth}\centering\scriptsize 14:00 \end{minipage} &
    \begin{minipage}{\customwidth}\centering\scriptsize 14:30 \end{minipage} &
    \begin{minipage}{\customwidth}\centering\scriptsize 15:00 \end{minipage} &
    \begin{minipage}{\customwidth}\centering\scriptsize 15:30 \end{minipage} &
    \begin{minipage}{\customwidth}\centering\scriptsize 16:00 \end{minipage} &
    \begin{minipage}{\customwidth}\centering\scriptsize 16:30 \end{minipage}
    \\
    \begin{sideways}\begin{minipage}{\customwidth}\centering \scriptsize 08/24/2013 \\ light clouds \vspace{5pt} \end{minipage}\end{sideways} &
    \includegraphics[width=\customwidth]{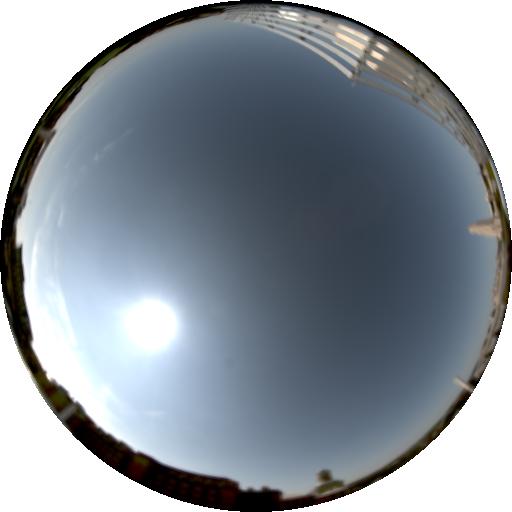} &
    \includegraphics[width=\customwidth]{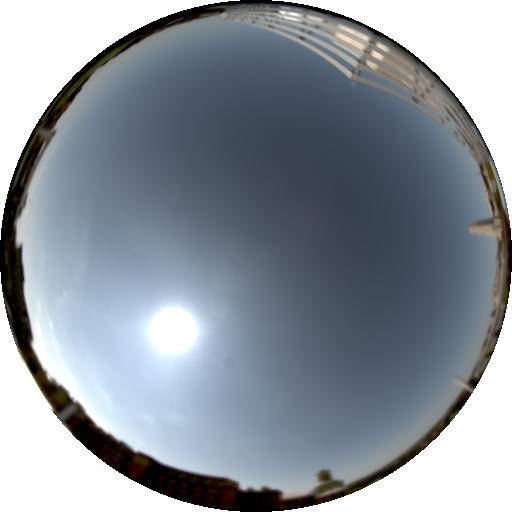} &
    \includegraphics[width=\customwidth]{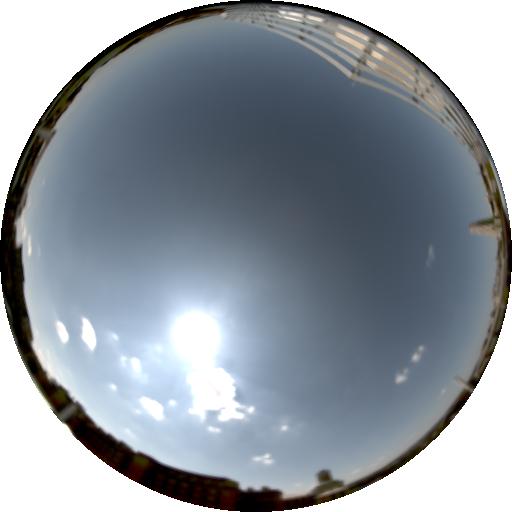} &
    \includegraphics[width=\customwidth]{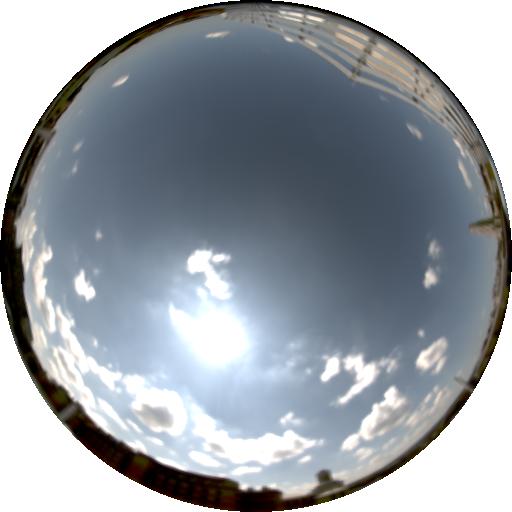} &
    \includegraphics[width=\customwidth]{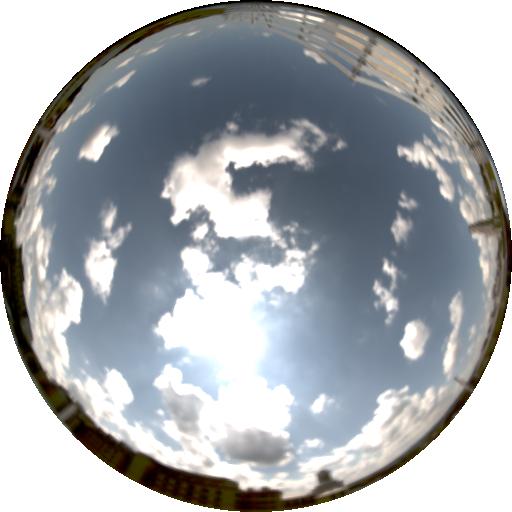} &
    \includegraphics[width=\customwidth]{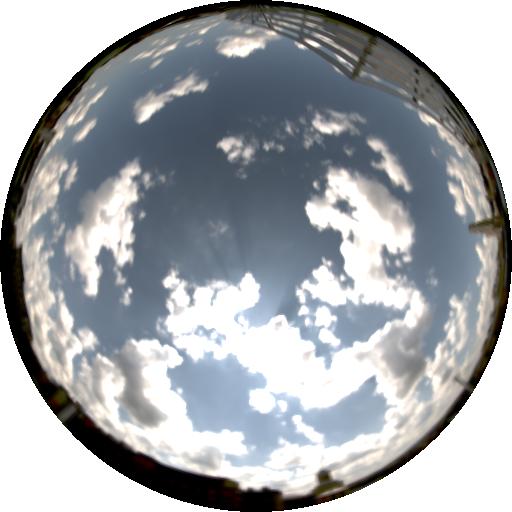} &
    \includegraphics[width=\customwidth]{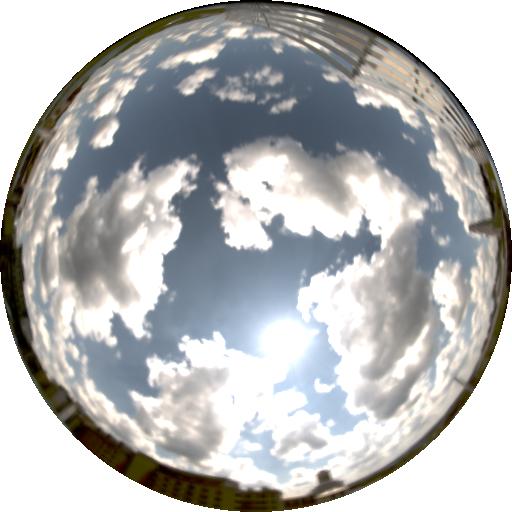} &
    \includegraphics[width=\customwidth]{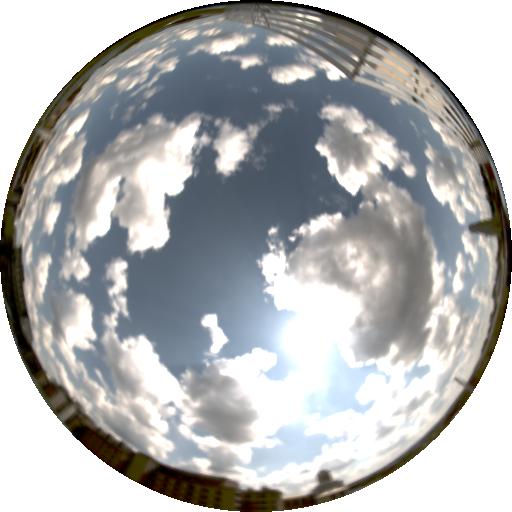} &
    \includegraphics[width=\customwidth]{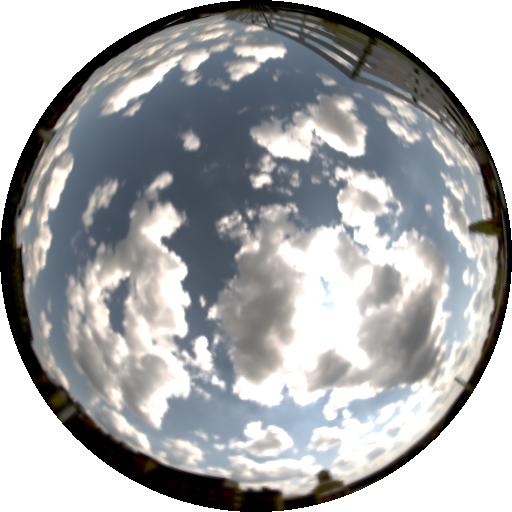} &
    \includegraphics[width=\customwidth]{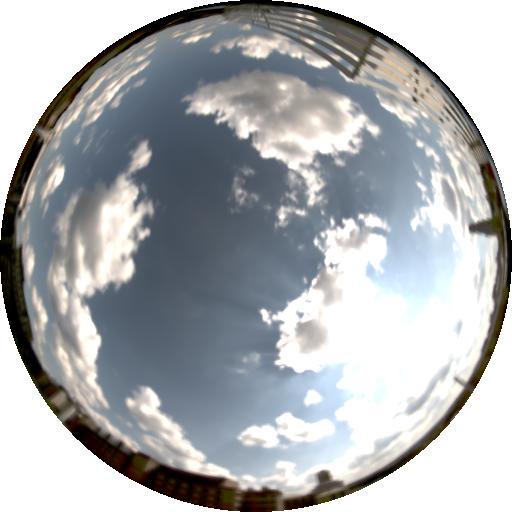} &
    \includegraphics[width=\customwidth]{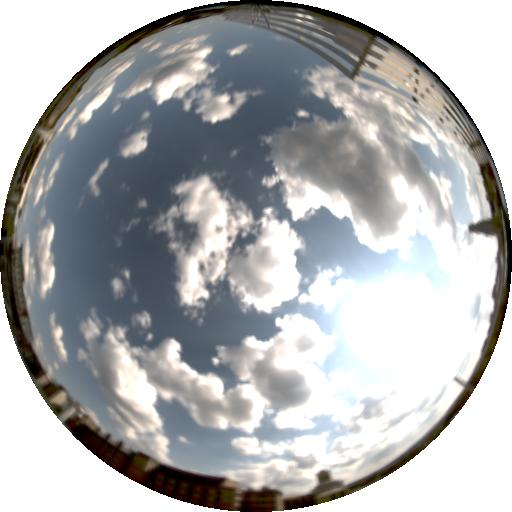} &
    \includegraphics[width=\customwidth]{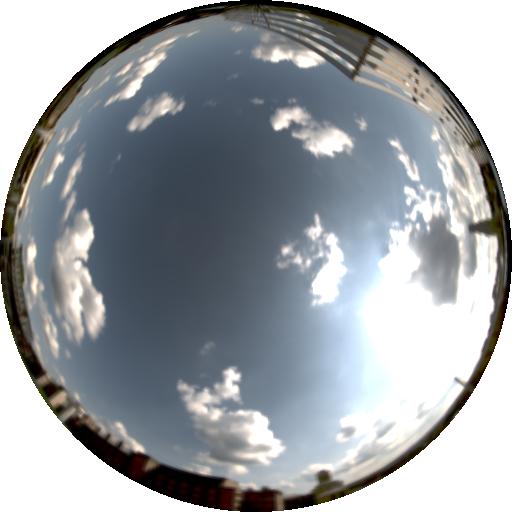}
    \\
    \begin{sideways}\begin{minipage}{\customwidth}\centering \scriptsize 11/06/2013 \\ mixed \vspace{5pt} \end{minipage}\end{sideways} &
    \includegraphics[width=\customwidth]{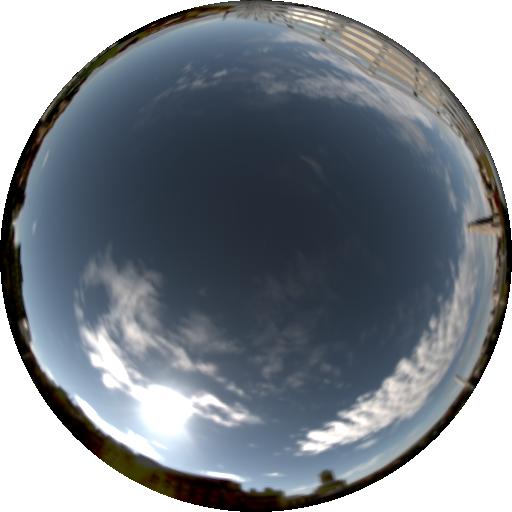} &
    \includegraphics[width=\customwidth]{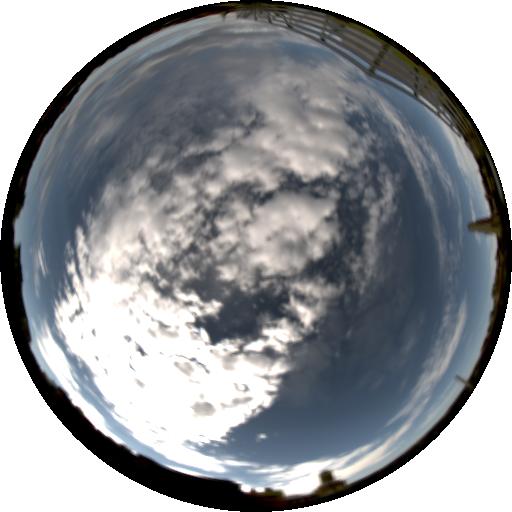} &
    \includegraphics[width=\customwidth]{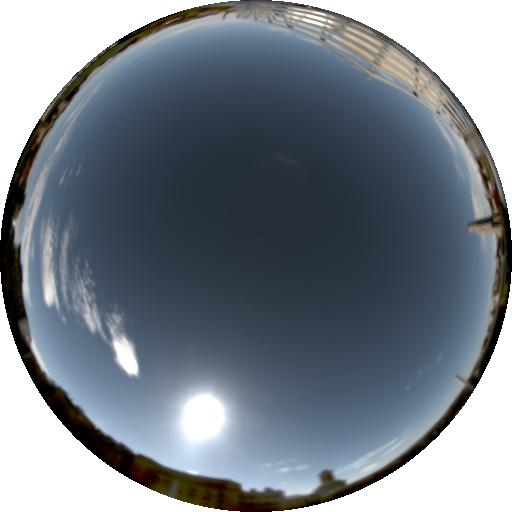} &
    \includegraphics[width=\customwidth]{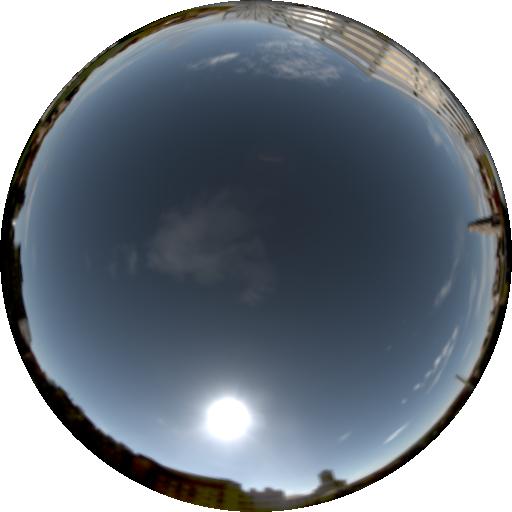} &
    \includegraphics[width=\customwidth]{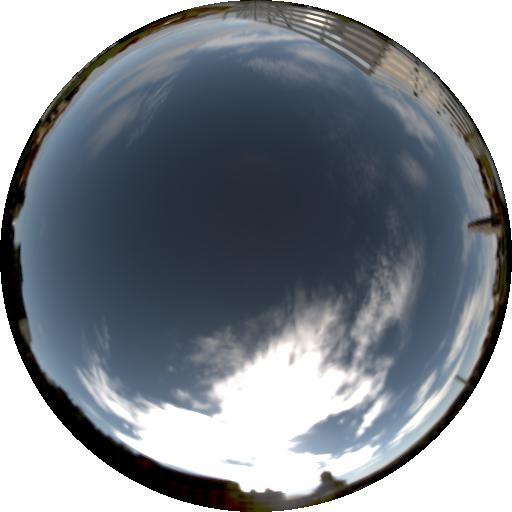} &
    \includegraphics[width=\customwidth]{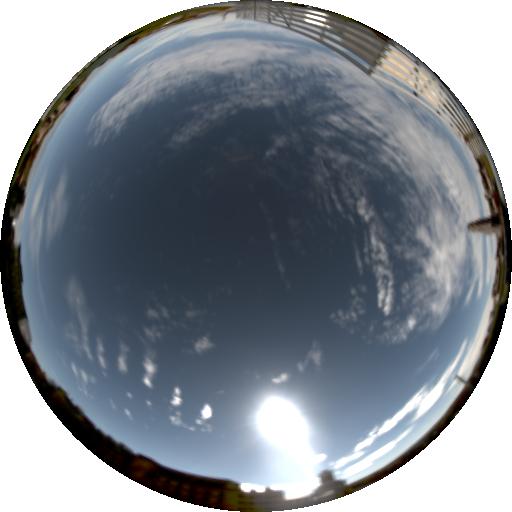} &
    \includegraphics[width=\customwidth]{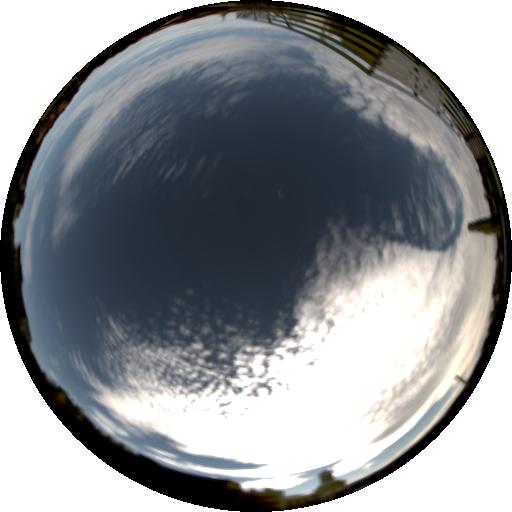} &
    \includegraphics[width=\customwidth]{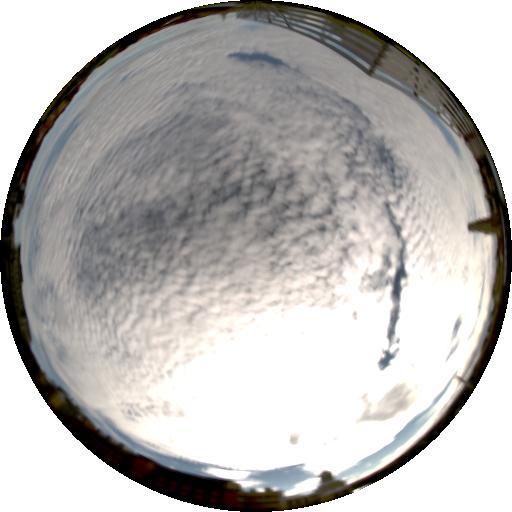} &
    \includegraphics[width=\customwidth]{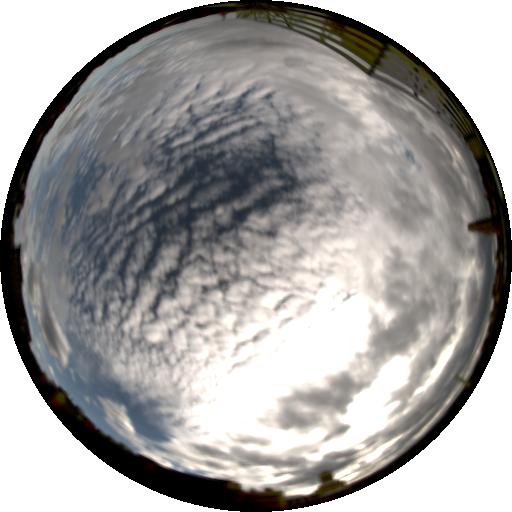} &
    \includegraphics[width=\customwidth]{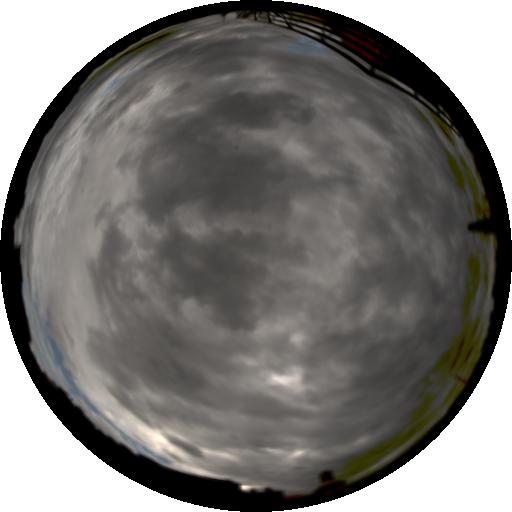} &
    \includegraphics[width=\customwidth]{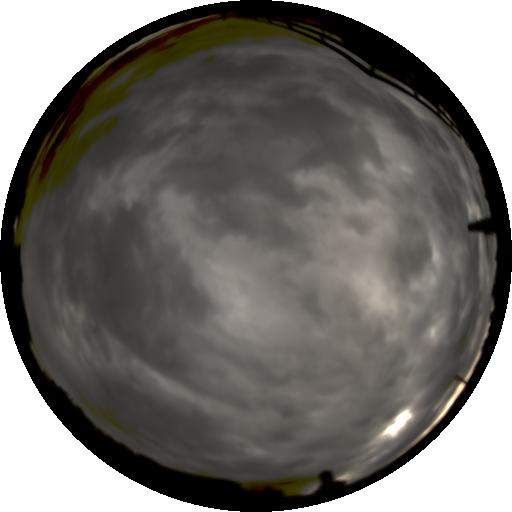} &
    \includegraphics[width=\customwidth]{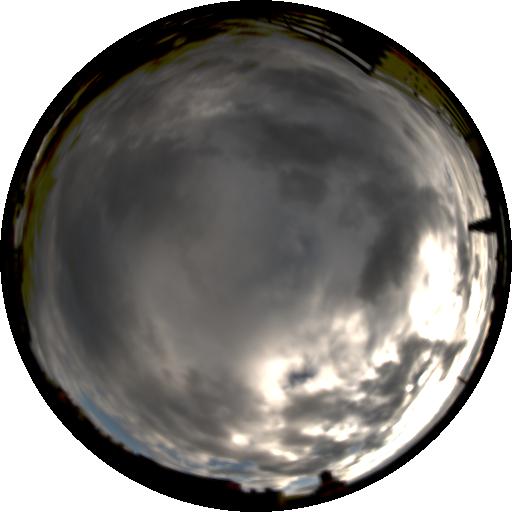}
    \\
    \begin{sideways}\begin{minipage}{\customwidth}\centering \scriptsize 11/08/2014 \\ heavy clds.\vspace{5pt} \end{minipage}\end{sideways} &
    \includegraphics[width=\customwidth]{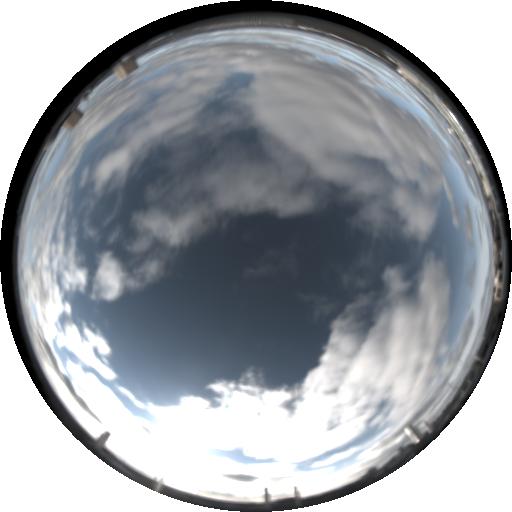} &
    \includegraphics[width=\customwidth]{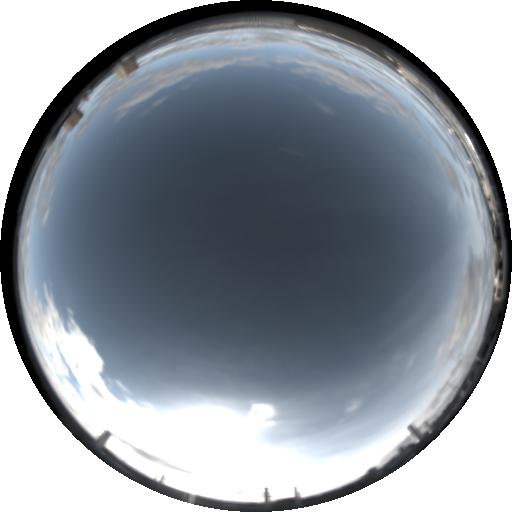} &
    \includegraphics[width=\customwidth]{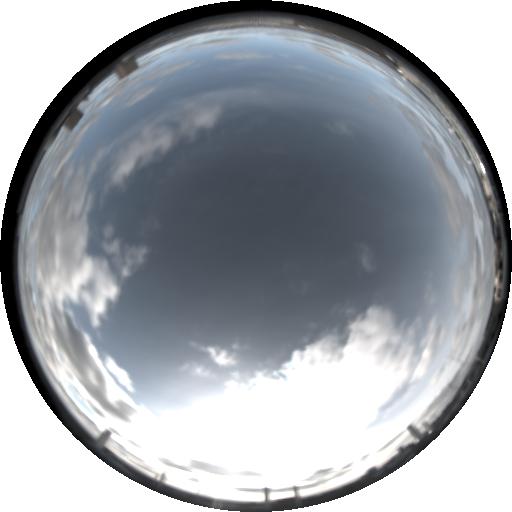} &
    \includegraphics[width=\customwidth]{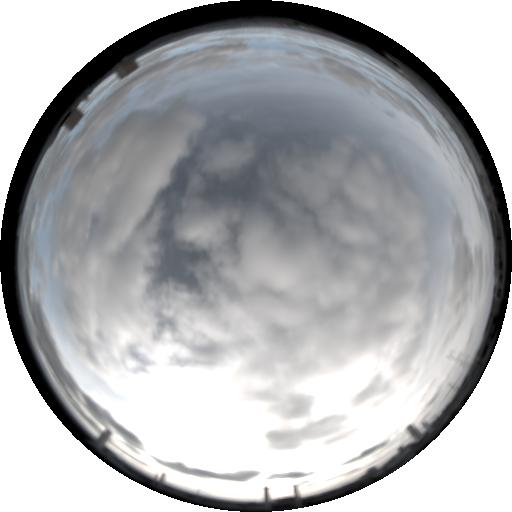} &
    \includegraphics[width=\customwidth]{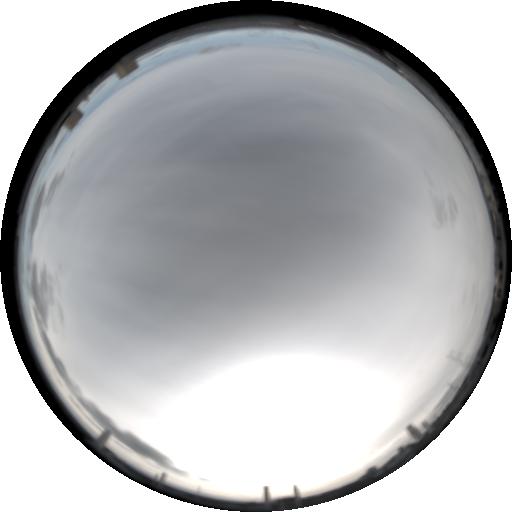} &
    \includegraphics[width=\customwidth]{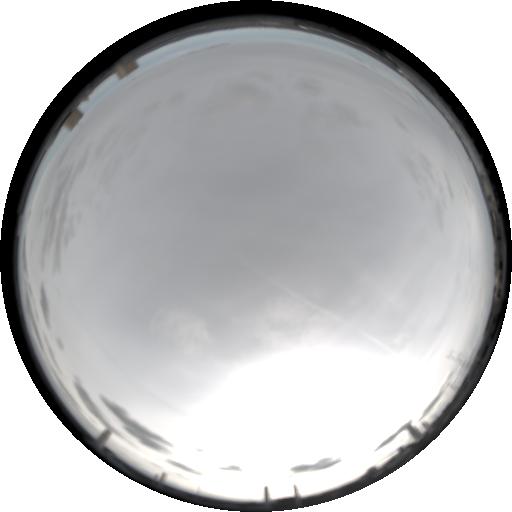} &
    \includegraphics[width=\customwidth]{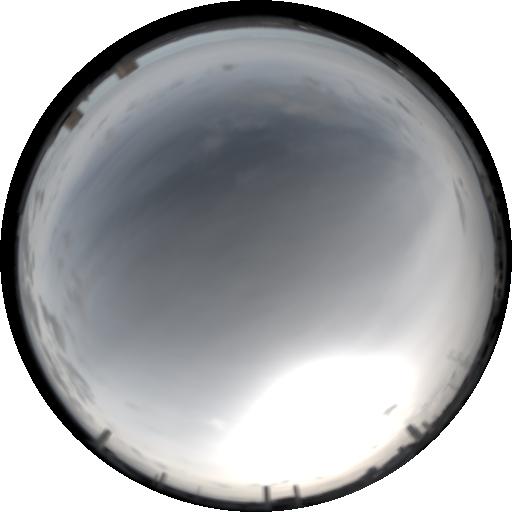} &
    \includegraphics[width=\customwidth]{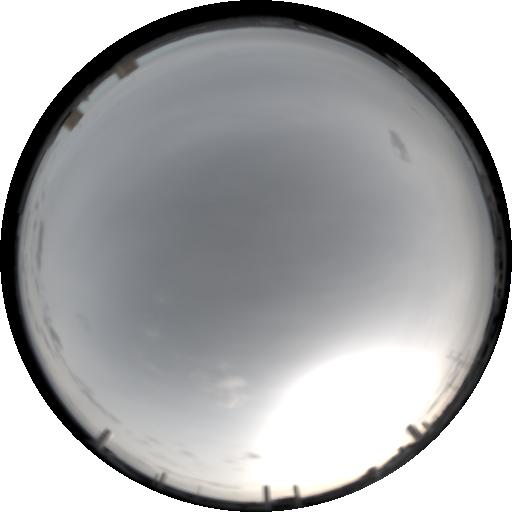} &
    \includegraphics[width=\customwidth]{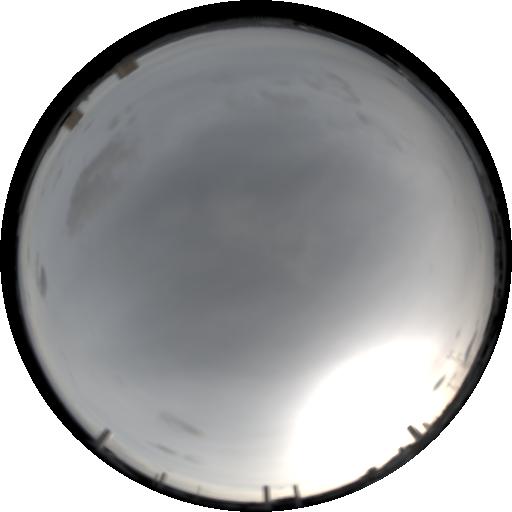} &
    \includegraphics[width=\customwidth]{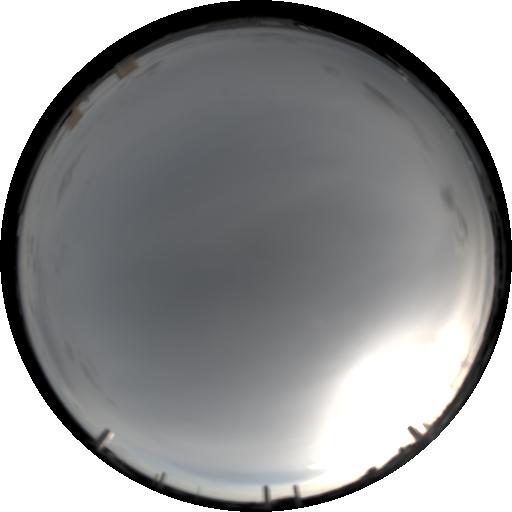} &
    \includegraphics[width=\customwidth]{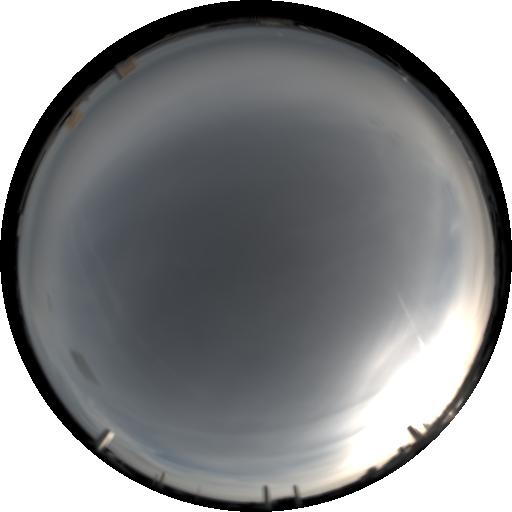} &
    \includegraphics[width=\customwidth]{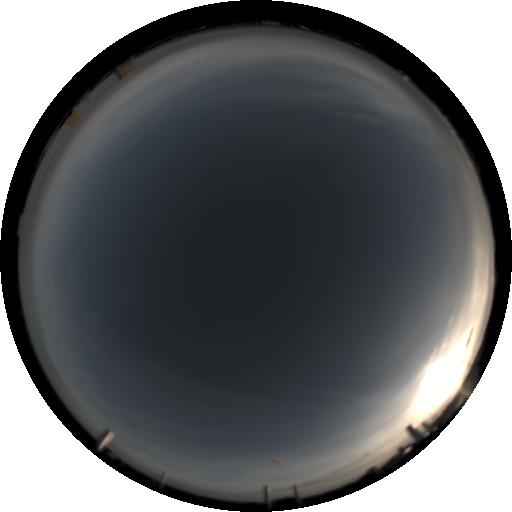}

    \\

    \end{tabular}
    \caption[HDRDB dataset excerpt]{Examples from our dataset of HDR outdoor illumination conditions. In all, our dataset contains 3,800 different illumination conditions, captured from 10{:}30 until 16{:}30, during 23 days, spread over ten months and at two geographical locations. Each image is stored in the 32-bit floating point EXR format, and shown tone mapped here for display (with $\gamma = 1.6$).}
    \label{fig:database}
\end{figure*}

\IEEEPARstart{S}{ince} its inception in the early 80s, Photometric Stereo~(PS)~\cite{woodham-opteng-80} has been explored under many an angle. Whether it has been to improve its ability to deal with complex materials~\cite{alldrin-cvpr-08} or lighting conditions~\cite{basri-ijcv-07}, the myriad of papers published on the topic are testament to the interest this technique has garnered in the community. While most of the papers on this topic have focused on images captured in the lab, recent progress has allowed the application of PS on images captured outdoors, lit by the more challenging case of uncontrollable, natural illumination. 



While capturing more data in the lab can be done relatively easily, the same cannot be said for outdoor imagery. Indeed, one does not control the sun and the other atmospheric elements in the sky; so one must wait for lighting conditions to change on their own. A creative solution to this problem was proposed in~\cite{hung-wacv-15}, but it is limited to objects that can be placed on a small moving platform. Therefore, capturing more data for fixed, large objects still means waiting days, or even months, potentially~\cite{ackermann-cvpr-12,abrams-eccv-12}.

A promising approach to answer this question is to use more elaborate models of illumination---high dynamic range (HDR) environment maps~\cite{reinhard-book-05}---as input to outdoor PS. Favorable results have been reported in~\cite{yu-iccp-13} for outdoor images taken in a single day, within an interval of just eight hours. However, the quality of outdoor results is reported to be inferior to that obtained in indoor environments. The decline is attributed to modest variation in sunlight, but no clear explanation is found in the literature. This observation leads to many interesting, unanswered questions: had the atmospheric conditions been different on that day, could the quality of their results have been better? Is a full day of observations enough to obtain good results in outdoor PS? 

This paper investigates PS in outdoor environments over the course of a single day and under a variety of sunlight conditions. 
Our first goal is to assess the \emph{reconstructibility} of surface patches as a function of their orientation and the illumination conditions. 
This is done using a large database of sky probes~\cite{hdrdb}, capturing the variability of natural, outdoor illumination. A detailed look at the conditions under which normals can be reconstructed reliably is presented, followed by an analysis of surface reconstruction stability. 

Our analysis reveals that reconstruction performance of classical PS methods can be categorized in two different sky types: partially cloudy and clear days. Interestingly, partially cloudy days typically offer better reconstruction accuracy, while clear days generally yield poor performance. During clear days, photometric cues alone do not provide a stable solution to the PS problem, leaving it under-constrained~\cite{woodham-opteng-80}. Our insight to solve this issue is to augment the photometric cues with learned features on geometry, reflectance and lighting to resolve ambiguities in singe-day outdoor PS.

We summarize our contributions as follows:
\begin{itemize}[noitemsep,nolistsep]
    \item an analysis of the conditioning of outdoor PS given photometric cues captured over a single day;
    \item a framework for predicting the performance of single-day outdoor PS with calibrated lighting, and its application in reconstructing surfaces on partially cloudy days;
	\item a state-of-the-art method for single-day outdoor PS with weakly-calibrated lighting, which is specifically designed to work on the ambiguous case of clear days. The method is robust to shadows, specularities and arbitrary but spatially uniform albedo.
\end{itemize}
Our contributions show that PS can be applied to images obtained over a single day under most weather conditions.

\section{Related work}

This section focuses on the more relevant work on outdoor PS, for conciseness. For an overview of general PS, the reader is referred to the recent, excellent review in~\cite{shi-tpami-18}.


Woodham's seminal work~\cite{woodham-opteng-80} shows that, for Lambertian surfaces, calibrated PS computes a (scaled) normal vector in closed form as a simple linear function of the input image pixels; this linear mapping is only well-defined for images obtained under three or more (known) non-coplanar lighting directions.
Subsequent work on outdoor PS has struggled to meet this requirement since, over the course of a day, the sun shines from directions that nearly lie on a plane. These co-planar sun directions then yield an ill-posed problem known as two-source PS; despite extensive research using integrability and smoothness constraints~\cite{onn-ijcv-90,hernandez-pami-11}, results still present strong regularization artifacts on surfaces that are not smooth everywhere. To avoid this issue in outdoor PS, authors initially proposed gathering months of data, watching the sun elevation change over the seasons~\cite{abrams-eccv-12,ackermann-cvpr-12}. Shen~{\em et al.}~\cite{shen-pg-14} noted that the intra-day coplanarity of sun directions actually varies throughout the year, with single-day outdoor PS becoming more ill-posed at high latitudes near the winter solstice, and worldwide near the equinoxes.

Another important issue is that, so far, most of the literature on PS has adopted a simple directional illumination model, for which optimal lighting configurations can be theoretically derived~\cite{drbohlav-iccv-05,klaudiny-prl-14,shen-pg-14}. Until recently, no attempt had been made to model natural lighting more realistically in an outdoor setup, where lighting cannot be controlled and atmospheric effects are difficult to predict. In such an uncontrolled environment, exploiting the subtlety and richness of natural lighting is key to improve the conditioning of PS and successfully apply it in the wild and with short intervals of time.


Thus, more recent approaches have attempted to compensate for limited sun motion by adopting richer illumination models that account for additional atmospheric factors in the sky. This is done by employing \mbox{(hemi-)spherical} environment maps~\cite{debevec-siggraph-98} that are either real sky images~\cite{yu-iccp-13,shi-3dv-14,hung-wacv-15,mo-cvpr-2018} or synthesized by parametric sky models~\cite{inose-tcva-13,jung-ijcv-19}. Despite these developments, state-of-the-art approaches in calibrated~\cite{yu-iccp-13} and semi-calibrated~\cite{jung-ijcv-19} (based on precise geolocation) outdoor PS are still prone to potentially long waits for ideal conditions to arise in the sky; and verifying the occurrence of such events is still a trial-and-error process.



Under more extreme ambiguity, techniques for shape-from-shading (SfS)~\cite{oxholm-eccv-12,johnson-cvpr-11,barron-pami-15} attempt to recover 3D normals from a single input image, in which case the shading cue alone is obviously insufficient to uniquely define a solution. Thus, SfS relies strongly on priors of different complexities and deep learning is quickly bringing advances to the field~\cite{eigen-iccv-15,shu-cvpr-17,wu-nips-17}. While this is encouraging, here we seek to improve the accuracy of 3D normal estimation by relying less heavily on priors and more strongly on the photometric cues obtained from multiple images. Finally, most of these methods are limited to a specific type of object and reflectance model ({\em e.g.}, human faces, Lambertian~\cite{shu-cvpr-17}).

\section{Overview}
\label{sec:overview}
\label{sec:hdrdb}

In this paper, we investigate the complex, natural lighting phenomena that help condition outdoor PS. Our analysis uses the Laval HDR Sky Database~\cite{hdrdb,lalonde-3dv-14}, a rich dataset of high dynamic range (HDR) images of the sky, captured under a wide variety of weather conditions. In all, the dataset totals more than 5,000 illumination conditions, captured over 50 different days. Fig.~\ref{fig:database} shows examples of these environment maps, which are tone mapped for display only; the actual sky images have a dynamic range that spans the full 22 stops required to properly capture outdoor lighting. 

Our investigations have approached outdoor PS under two different scenarios, leading to two specialized solutions:

\begin{enumerate}[noitemsep,leftmargin=12pt,topsep=1pt]
\item First, we consider the popular case of calibrated, outdoor PS for Lambertian objects (sec.~\ref{sec:LambertianCalibrated}) and we assess how outdoor PS is conditioned solely by the few photometric cues obtained over the course of one day. By considering Lambertian reflectance, the number of unknowns is reduced to a minimum and, therefore, this analysis provides an upper bound on the quality of recovered normals for objects with general reflectances. As we initially reported in~\cite{holdgeoffroy-iccp-15,holdgeoffroy-3dv-15}, partly cloudy days are in fact better for single-day outdoor PS since clouds obscure and further scatter sun light, causing a beneficial shift in the effective direction of illumination. Such conditions lend themselves well to calibrated PS algorithms. On the other hand, our analysis also suggest that a different approach is needed for outdoor PS with clear skies and objects with more general reflectances.
\item Second, we consider non-Lambertian objects and the more difficult, under-constrained case of sunny days with clear skies  (sec.~\ref{sec:NonLambertianSemiCalibrated}). In addition, we also relax the assumption on fully-calibrated lighting. Since there are more unknowns in this new scenario, we cannot rely solely on the few photometric cues obtained within a single day. We thus propose an approach that uses deep learning to resolve ambiguities in outdoor PS by aggregating prior knowledge on object geometry, material and their interaction with natural outdoor illumination. This new approach is the first of its kind---so far, deep PS had only been applied in indoor scenarios with rich and controlled illumination~\cite{yu-iccv-17,santo-iccv-17,taniai-arxiv-18,shi-tpami-18}.
\end{enumerate}
We conclude by discussing how the advantages of the two solutions above could be integrated into a single, more generic approach in future work.

\section{Lambertian, Calibrated Outdoor PS} \label{sec:LambertianCalibrated}
\subsection{Image formation model}
\label{sec:ifm}

\begin{figure}[t]
    \centering
    \includegraphics[width=.42\linewidth]{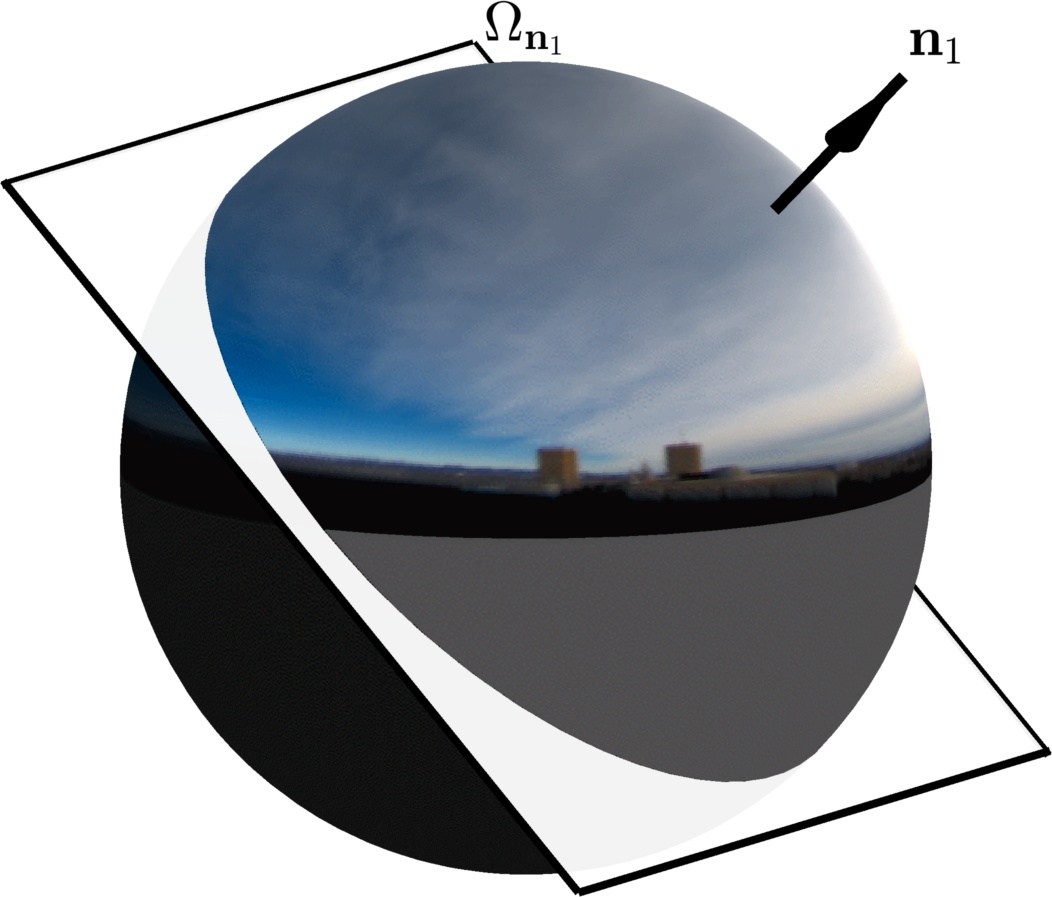}
    \includegraphics[width=.42\linewidth]{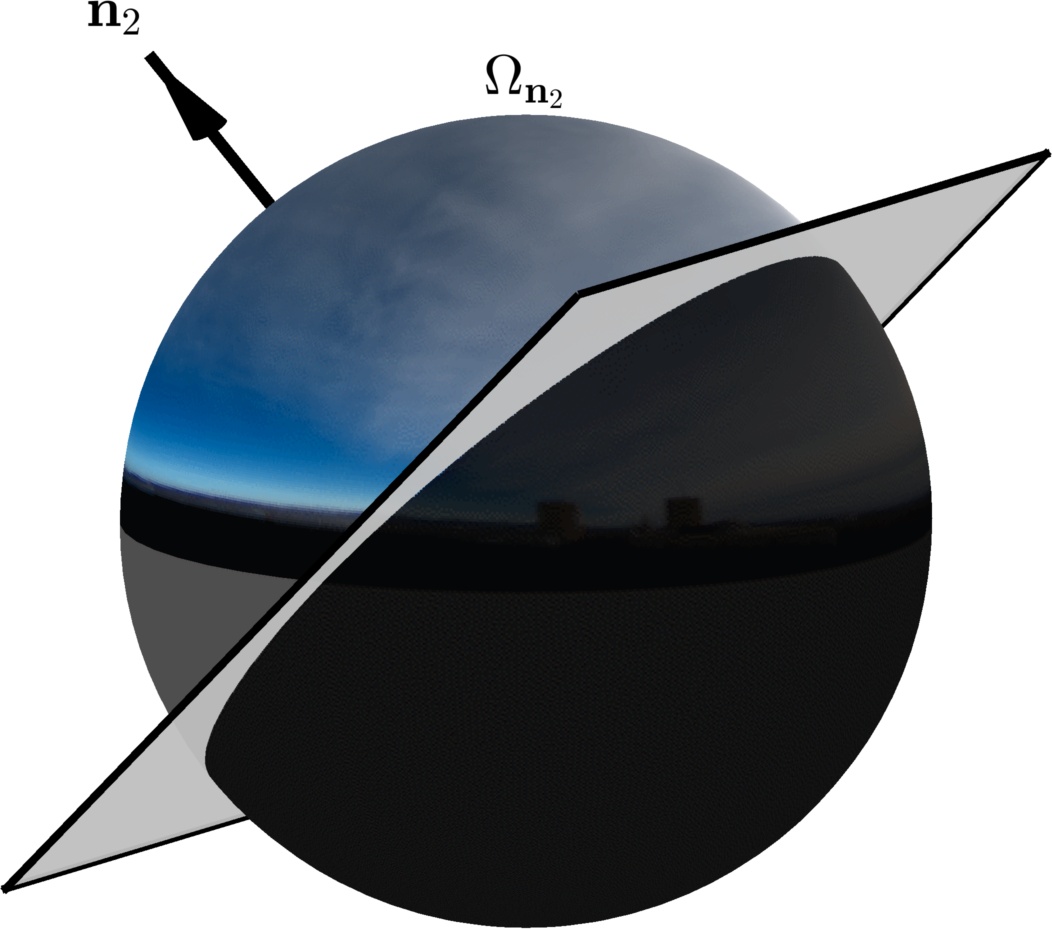}
    \caption[Hemispheres defined by surface normals]{A normal $\mathbf{n}$ defines an integration hemisphere $\Omega_{\mathbf{n}}$ on the environment map. Only light emanating from this hemisphere contributes to the shading on that patch. Thus, patches with different normals are lit differently even if the environment map is the same.}
    \label{fig:normal-diagram}
\end{figure}

Consider a small, Lambertian surface patch with normal vector ${\bf n}$ and albedo $\rho$ (w.l.o.g., assume albedo is monochromatic). At time $t$, this surface patch is observed under natural, outdoor illumination represented by the environment map $L_t(\boldomega)$ (\eg, fig.~\ref{fig:database}), with $\boldomega$ denoting a direction in the unit sphere. With an orthographic camera model, this patch is depicted as an image pixel with intensity
\begin{equation}
    b_t = \frac{\rho}{\pi} \int_{\Omega_{\bf n}} {\bf L}_t(\mathbf{\boldomega}) \langle \boldomega, {\bf n} \rangle d\boldomega \,,
    \label{eqn:imageformation-continuous}
\end{equation}
where $\langle \cdot, \cdot \rangle$ denotes the dot product. Integration is carried out over the hemisphere of incoming light, $\Omega_{\bf n}$, defined by the local orientation ${\bf n}$ of the surface, fig.~\ref{fig:normal-diagram}. This hemisphere corresponds to an occlusion (or attached shadow) mask; only half of the pixels in the environment map contribute to the illumination of the surface patch. To make the analysis tractable and independent of object geometry, this analysis focuses on the simpler case without cast shadows.

This image formation model is then discretized as,
\begin{equation}
b_t = \frac{\rho}{\pi}\sum_{\boldomega_j \in \Omega_{\bf n}} \hat{\bf L}_t(\boldomega_j) \langle \boldomega_j, {\bf n} \rangle\,,
\label{eqn:imageformation-discrete}
\end{equation}
with $\hat{\bf L}_t(\boldomega_j) = {\bf L}_t(\boldomega_j)\Delta\boldomega_j$ representing the environment map weighted by the solid angle $\Delta\boldomega_j$ spanned by pixel $j$ ($\Delta\boldomega_j$, $\forall j$, are normalized as to sum to $2\pi$). Eq.~$\eqref{eqn:imageformation-discrete}$ can be further summarized into the equivalent form
\begin{equation}
b_t = {\bf \bar l}_t^T \mathbf{x}
\label{eqn:imageformation-simplified}
\end{equation}
where ${\bf x} = \rho {\bf n}$ is the albedo-scaled normal vector and
\begin{equation}
{\bf \bar l}_t = \frac{1}{\pi} \sum_{\boldomega_j \in \Omega_{\bf n}} \hat{\bf L}_t(\boldomega_j) \boldomega_j \in \mathbb{R}^3 \quad.
\label{eqn:mean-light}
\end{equation}
This vector ${\bf \bar l}_t$ can be interpreted as a virtual point light source summarizing the illumination provided by the environment map at a time $t$. This vector ${\bf \bar l}_t$ is the mean of the light vectors computed over the hemisphere of incoming light directions defined by ${\bf n}$ (see fig.~\ref{fig:mlv}). As such, this vector is henceforth referred to as the {\em mean light vector} (MLV). It is important to note that, as opposed to the traditional PS scenario where point light sources are fixed and thus independent of ${\bf n}$, here the {\em per-pixel} MLV is a function of $\mathbf{n}$. Thus, patches with different orientations define different sets of MLVs (see fig.~\ref{fig:normal-diagram}). A similar lighting representation has been adopted in the uncalibrated case in~\cite{mo-cvpr-2018}.


Given multiple images taken at times $t \in \{1,2,\ldots,T\}$, we collect all photometric constraints for patch ${\bf x}$ to obtain:
\begin{equation}
\mathbf{b} =
\begin{bmatrix}
 b_1 \\ b_2 \\ \vdots \\ b_T
\end{bmatrix}
=
\begin{bmatrix}
 {\bf \bar l}_1^T \\ {\bf \bar l}_2^T \\ \vdots \\ {\bf \bar l}_T^T
\end{bmatrix}
{\bf x} = \mathbf{L} \mathbf{x} \,.
\label{eqn:matrix-form}
\end{equation}
With eq.~\eqref{eqn:matrix-form}, this model of natural environmental illumination becomes quite similar to a model with a distant point light source, the well-known case in PS. However, note that each ${\bf \bar l}_t$ in ${\bf L}$ is a function of $\Omega_{\bf n}$ and, thus, of ${\bf n}$.

Most importantly, in outdoor PS, a well-defined solution ${\bf x}$ may exist even if the relative sun motion is nearly planar during a certain time interval. Instead of relying solely on sun direction, now, the solution requires non-coplanar mean light vectors ${\bf \bar l}_t$, which are determined by a comprehensive set of natural illumination factors.

\subsection{Measuring uncertainty}
\label{sec:modeling_uncertainty}

From eq.~\eqref{eqn:matrix-form}, the least-squares solution ${\bf x} = ({\bf L}^T{\bf L})^{-1}{\bf L}^T{\bf b}$ of outdoor PS is clearly affected by the condition number of ${\bf L}$. Thus, we next characterize how well the solution ${\bf x}$ is constrained by natural, outdoor illumination within a given time interval (\eg, one day)---which is encoded by the set of mean light vectors ${\bf \bar l}_t$ in ${\bf L}$ or, equivalently, the set of environment maps $L_t(\cdot)$.

To assess the reliability of a solution ${\bf x}$, we follow standard practice in PS~\cite{klaudiny-prl-14,sun-ivc-07} and consider image measurements corrupted by zero-mean Gaussian noise with equal variance $\sigma^2$ (as least squares estimation is only optimal for this practical, most common noise model). Thus, ${\bf b}$ in eq.~\eqref{eqn:matrix-form} follows a normal distribution:
\begin{equation}
{\bf b} \sim \mathcal{N}\left( \boldmu_b, \sigma^2 {\bf I} \right)\,,
\end{equation}
where $\boldmu_b$ has the (unknown) uncorrupted pixel values.

Since the desired least-squares solution for the albedo-scaled normal, ${\bf x} = \left({\bf L}^T{\bf L}\right)^{-1}{\bf L}^T{\bf b}$, is a linear transformation of a Gaussian random vector, it is easy to show that
\begin{equation}
{\bf x} \sim \mathcal{N} \left( \boldmu_x, \sigma^2({\bf L}^T{\bf L})^{-1} \right)\,,
\end{equation}
where $\boldmu_x = \left({\bf L}^T{\bf L}\right)^{-1}{\bf L}^T\boldmu_b$ is the expected value of ${\bf x}$.
Once the albedo of a surface patch is known, we analyze its contribution to the uncertainty in the estimated normal vector, ${\bf n} = \rho^{-1}{\bf x}$, using a similar distribution,
\begin{equation}
{\bf n} \sim \mathcal{N} \left( \frac{\boldmu_x}{\rho}, \frac{\sigma^2}{\rho^2}({\bf L}^T{\bf L})^{-1} \right)\,.
\label{eqn:normal-distrib}
\end{equation}

The marginal distributions in eq.~\eqref{eqn:normal-distrib} allow us to derive confidence intervals that indicate the uncertainty in each component of the least squares estimate ${\bf \hat n} = [ \hat n_x \ \hat n_y \ \hat n_z ]^T$ of ${\bf n} = [ n_x \ n_y \ n_z ]^T$. The corresponding $95\%$ confidence interval~\cite{hastie-book-09} is given by
\begin{equation}
\hat{\mathbf{n}} \pm \bolddelta \,, \quad \text{with } \delta_k = 1.96 \frac{\sigma\lambda_k}{\rho} \,,
\label{eqn:confidence-xyz}
\end{equation}
where $\lambda_k$ is the square root of the $k$th element on the diagonal of $({\bf L}^T{\bf L})^{-1}$. As expected, the sensor-dependent noise level $\sigma$ is not the only factor that determines uncertainty. The noise gain factor $\lambda_k$ in eq.~\eqref{eqn:confidence-xyz} reveals how outdoor illumination (the conditioning of ${\bf L}$) can amplify the effect of noise on the solution ${\bf \hat n}$. The albedo $\rho$ also impacts the solution stability, where a lower albedo translates in a larger variance in the obtained estimate $\bf \hat n$ (as less light is reflected towards the camera). Our goal is then to answer the remaining question: how do natural changes in outdoor illumination affect this noise gain factor ($\lambda_k$) and, therefore, uncertainty?


To provide a measure of uncertainty that is more intuitive than eq.~\eqref{eqn:confidence-xyz}, we consider angular distances in degrees,
\begin{align}
\theta^\pm &= \cos^{-1}(\mathbf{n}^T{\bf \hat n}^\pm)\,, \qquad
{\bf \hat n}^\pm = \frac{{\bf \hat n} \pm \bolddelta}{\lVert{\bf \hat n} \pm \bolddelta \rVert}\,.
\label{eqn:angular-dist}
\end{align}
The uncertainty in the estimate of ${\bf n}$ is then summarized in a single confidence interval, in degrees,
\begin{equation}
\mathcal{C}_{\bf n} = [ \ 0, \ \max (\theta^\pm) \ ]\,,
\label{eqn:confidence-degrees}
\end{equation}
which indicates the expected accuracy of the estimated surface orientation ${\bf \hat n}$.

Note that the condition number, determinant, and trace of matrix $(\mathbf{L}^T\mathbf{L})^{-1}$ can also be used as measures of total variance in the estimated solutions---as done in~\cite{sun-ivc-07}---to find the optimal location of point light sources in PS. These measures are closely related to the rank of matrix $\mathbf{L}$, which must be three for a solution to exist; that is, $\mathbf{L}^T\mathbf{L}$ must be nonsingular. In practice, this matrix is always full-rank, although it is often poorly conditioned~\cite{shen-pg-14}. In the following, we also consider the gain factor $\lambda_k$ in~\eqref{eqn:confidence-xyz} as a measure of uncertainty independent of albedo and sensor noise. We focus on analyzing our ability to recover geometry and will assume that the albedo is constant.

\begin{figure}[t]
    \centering
    \includegraphics[width=.6\linewidth]{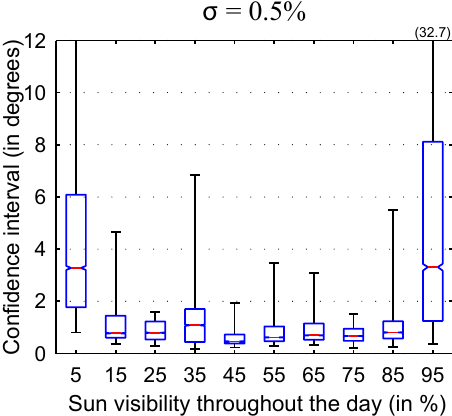}
    \vspace{-2mm}
    \caption[Confidence interval of normal estimates in function of sun visibility]{Median confidence interval of normal estimates (red line) as a function of mean sun visibility over the course of the day for a signal noise $\sigma = 0.5\%$, in bins of $10^{\circ}$. Our analysis predicts that normal reconstruction errors will likely be high if the sky is completely overcast (low sun visibility), or completely clear (high sun visibility). Good results can thus be expected in partially cloudy conditions, as shown in fig.~\ref{fig:cloud-cover}. The lower (upper) edge of each blue box indicates the 25th (75th) percentile. Statistics are computed only on normals pointing upwards to lessen ground effects.}
    \label{fig:cloud-cover-plot}
\end{figure}

\subsection{Effect of clouds on outdoor lighting}
\label{sec:cloud-cover-results}

This section investigates the environmental element that most influences mean light vectors throughout the day: {\em clouds}.
Cloud coverage has an important effect on the uncertainty of normal reconstruction because clouds introduce variability in illumination and, thus, new photometric cues as they (dis)occlude the sun. Here, we present a systematic analysis of their influence on outdoor PS. To control for the effect of the sun elevation, the analysis is performed on 23 days with similar sun elevations by keeping only the skies captured in October and November.

We approximate cloud coverage by computing the fraction of time that the sun is visible, \ie, that it fully shines on the scene, for a given day. To do so, we simply find the brightest spot in a sky image, and determine that the sun shines on the scene if the intensity of the brightest pixel is greater than $20\%$ of the maximum sun intensity---we determined empirically that this is the point at which the sun is bright enough to start creating cast shadows. Cloud coverage is represented by computing the mean sun visibility for a given day. Values of less than 10--15\% indicate mostly overcast skies, while skies are mostly clear if above 85--90\%.

\begin{figure}
 \centering
 \begin{minipage}{\linewidth}
 \centering
 \begin{sideways}\begin{minipage}{.2\linewidth}\centering \scriptsize Clear (85-100\%)\vspace{0pt} \end{minipage}\end{sideways}
 \includegraphics[width=.425\linewidth]{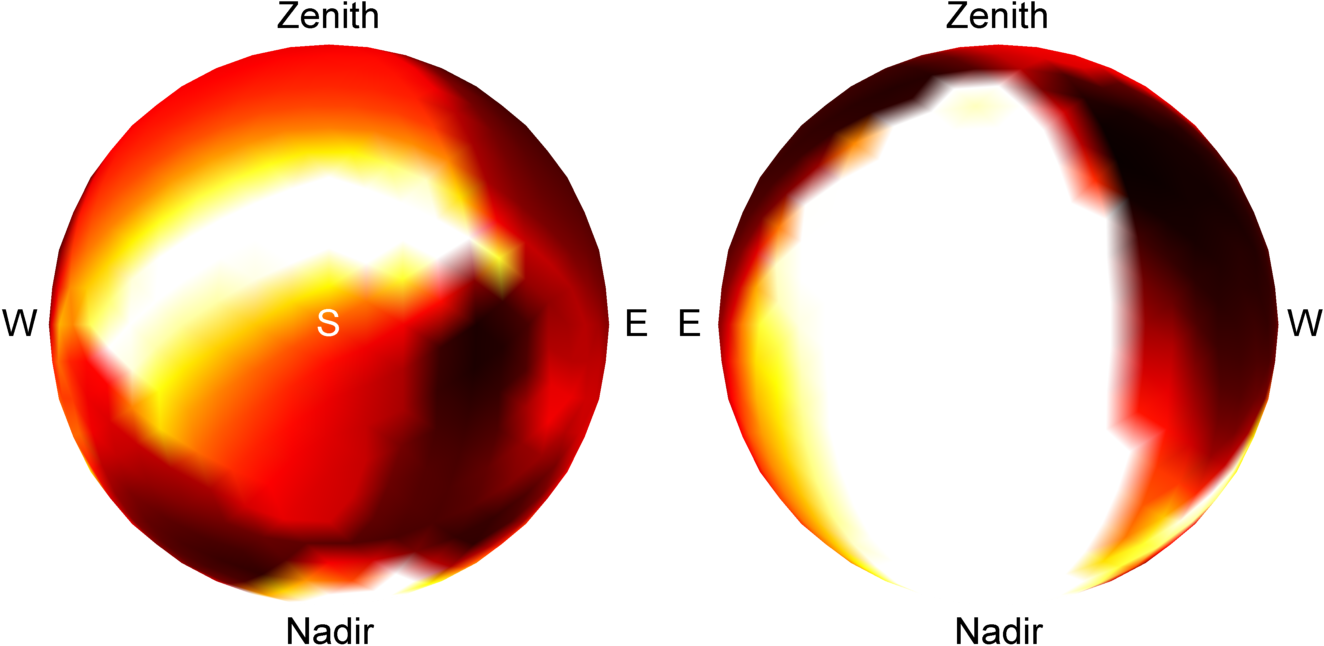} 
 \begin{sideways}\begin{minipage}{.2\linewidth}\centering \scriptsize Mixed Clear (50-85\%)\vspace{0pt} \end{minipage}\end{sideways}
 \includegraphics[width=.425\linewidth]{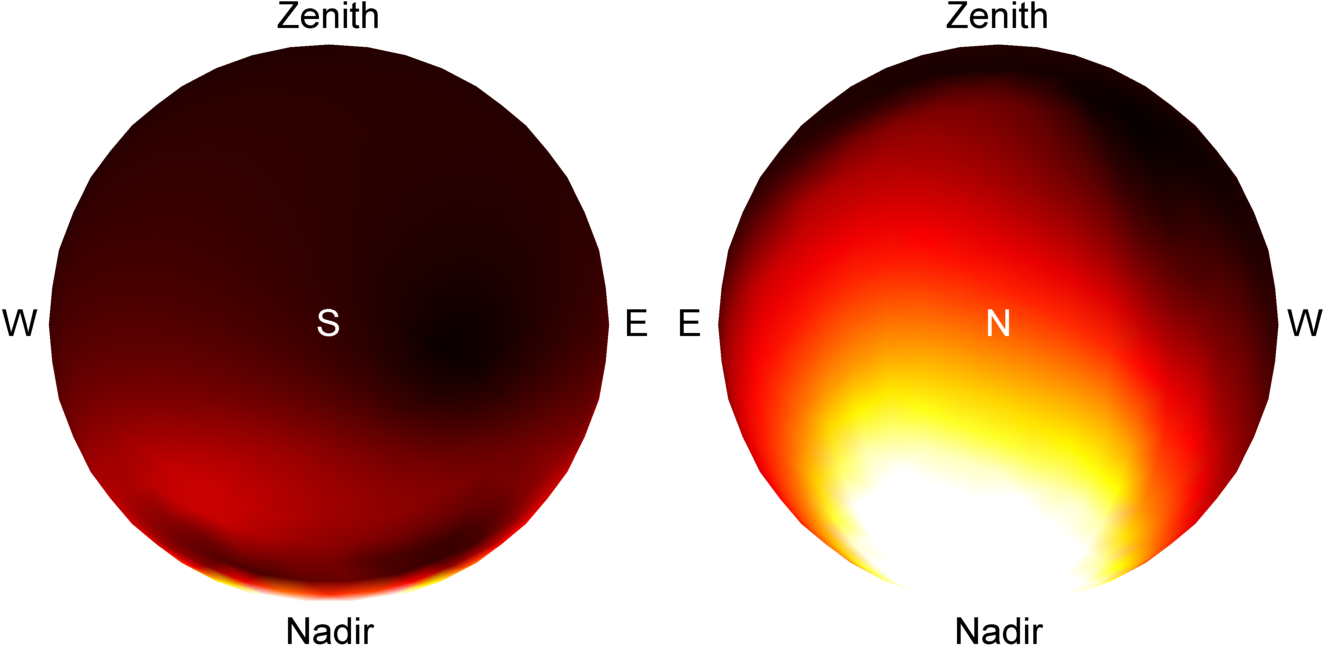} \\
 \noindent\rule{\linewidth}{0.1pt} 
 \begin{sideways}\begin{minipage}{.225\linewidth}\centering \scriptsize Mixed (15-50\%)\vspace{0pt} \end{minipage}\end{sideways}
 \includegraphics[width=.425\linewidth]{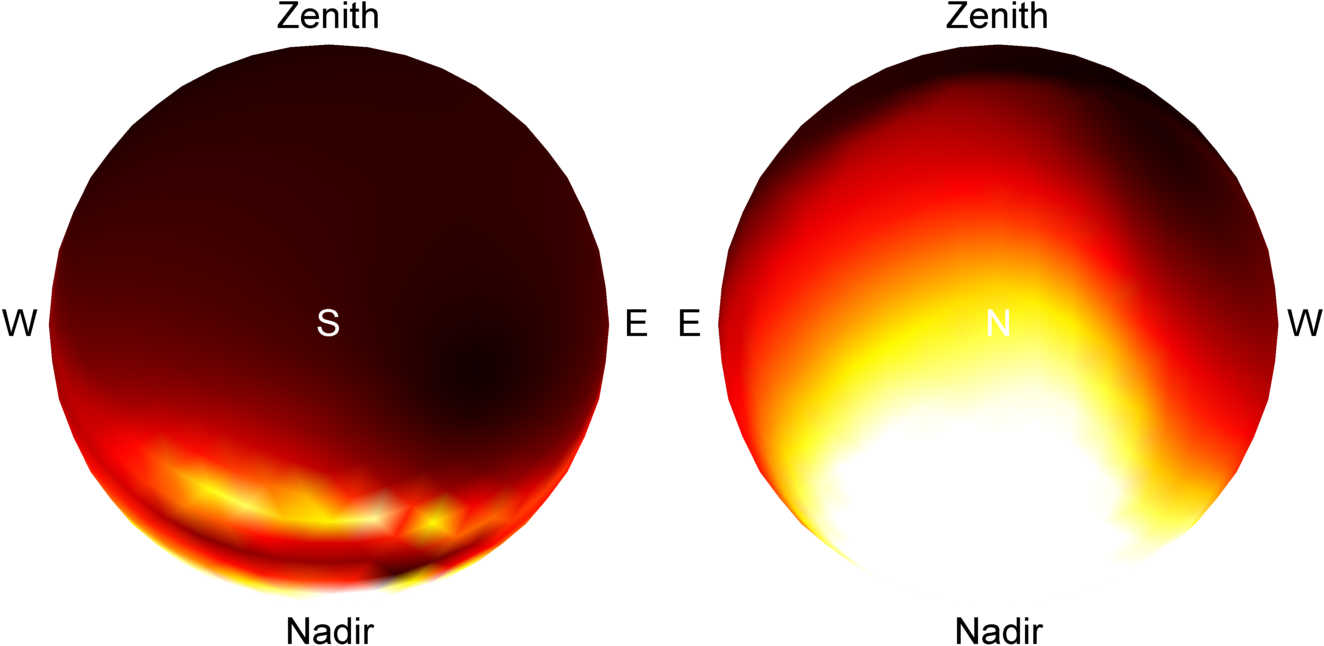} 
 \begin{sideways}\begin{minipage}{.2\linewidth}\centering \scriptsize Overcast (0-15\%)\vspace{0pt} \end{minipage}\end{sideways}
 \includegraphics[width=.425\linewidth]{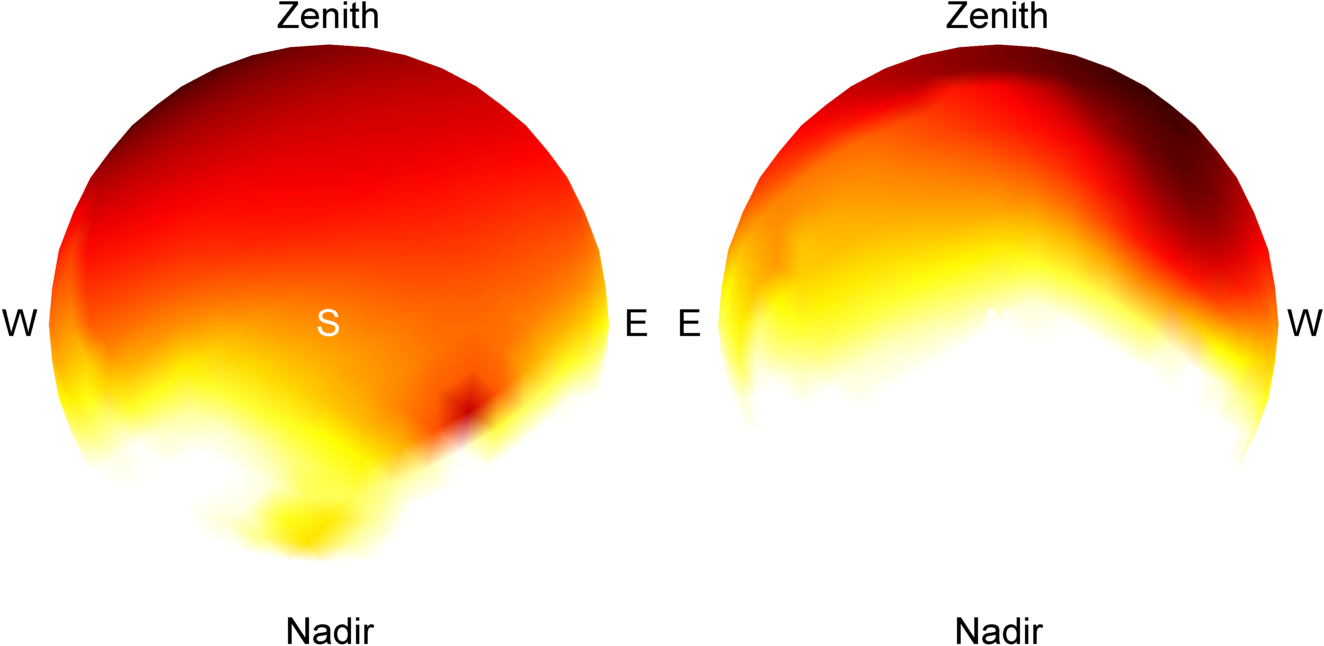} \vspace{0.8em}\\
 \end{minipage}
 \includegraphics[width=0.60\linewidth,height=11pt]{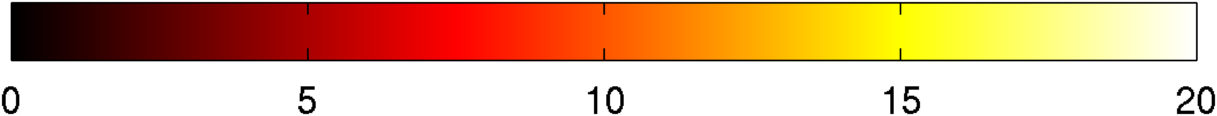} 
 \caption{Influence of cloud cover on the 95\% confidence intervals (in degrees) with $\sigma=1\%$. Each pair of plots show the full sphere of normals from two different viewpoints: South (left), and North (right). Four different types of skies are shown, based on sun visibility. For example, the top-left plots show the confidence intervals averaged over all days with direct sun visibility in the range 85\%-100\%. }
 \label{fig:cloud-cover}
\end{figure}

The relation between sun visibility and the confidence interval $\mathcal{C}_\mathbf{n}$ is shown in fig.~\ref{fig:cloud-cover-plot}. Photometric-related normal reconstruction errors will likely be quite high in two situations: completely overcast (low sun visibility), or completely clear skies (high sun visibility). Interestingly, good reconstruction results are expected for a wide range of cloud coverage conditions, ranging from 10--90\% mean sun visibility.

These results are corroborated by fig.~\ref{fig:cloud-cover}, which shows the confidence intervals themselves.  
These intervals are averages over skies belonging to four groups: overcast (0--15\%), mixed overcast (15--50\%), mixed clear (50--85\%), and clear (85--100\%) days. Again, high uncertainty results are visible for the two extreme cases of fully overcast and fully clear days, while the remainder indicate more stable solutions.

\begin{figure}[t]
\centering
\includegraphics[width=\columnwidth]{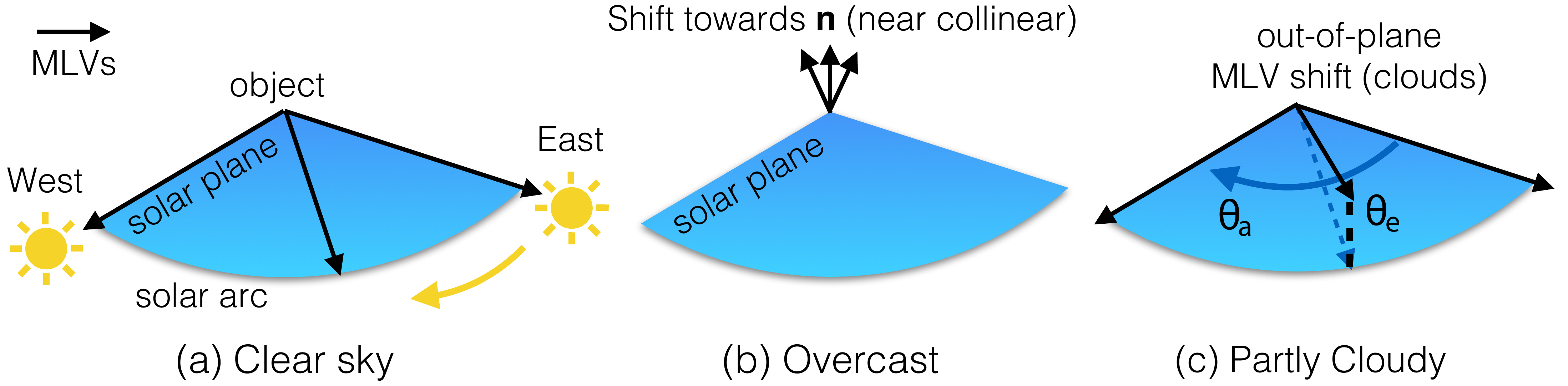}
\caption[Impact of cloud coverage on PS conditioning]{Impact of cloud coverage on the numerical conditioning of outdoor PS: clear (a) and overcast (b) days present MLVs with stronger coplanarity; in partly cloudy days (c) the sun is often obscured by clouds, which may lead to out-of-plane shifts of MLVs.}
\label{fig:MLVshift}
\end{figure}

\begin{figure}
 \centering
 \includegraphics[width=0.85\linewidth]{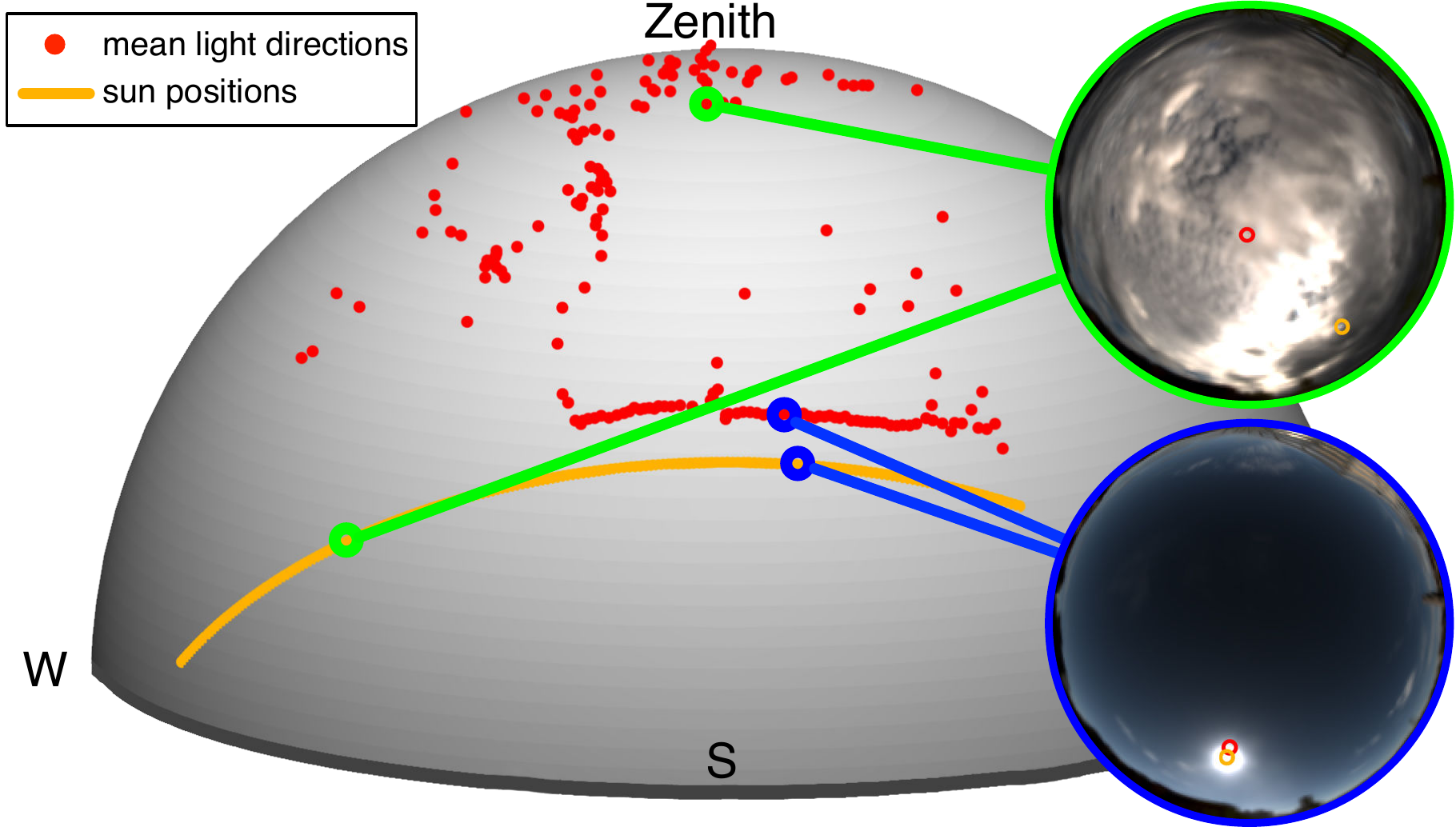}
 \caption[Cloud effect on mean light direction]{Cloud effect on the MLV over one day: while the sun path (orange) yields nearly co-planar directions of illumination, the mean light directions (red dots) for a normal pointing up provide a much more varied set (data from 11/06/2013, second row of fig.~\ref{fig:database}).}
    \label{fig:mlv}
\end{figure}

Under clear skies, the MLVs ${\bf \bar l}_t$ of the model above will point nearly towards the sun, from which arrives most of the incoming light. Thus, near an equinox (worldwide), the associated MLVs are nearly coplanar~\cite{shen-pg-14}, resulting in poor performance, fig.~\ref{fig:MLVshift}~(a). For a day with an overcast sky, performance is also poor because the set of MLVs are nearly colinear and shifted towards the patch normal ${\bf n}$, fig.~\ref{fig:MLVshift}~(b).
The improved conditioning in mixed skies is explained by the following key observation: cloud cover \emph{shifts} the MLVs ${\bf \bar l}_t$ towards zenith and away from sun trajectory in the sky, fig.~\ref{fig:MLVshift}~(c). Therefore, even when the sun moves along a trajectory that nearly lies on a 3D plane, as shown in fig.~\ref{fig:mlv}, cloud cover effectively causes an {\em out-of-plane} shift of the MLVs, making reconstruction possible.


\begin{figure}[t]
\centering
\hspace{3.1cm} \includegraphics[width=2.6cm]{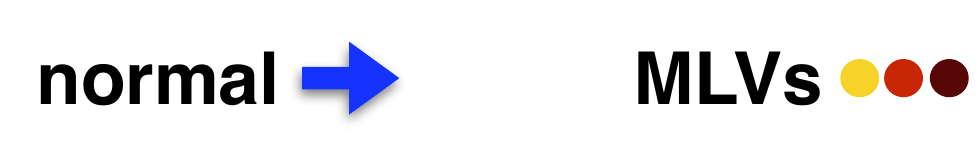} \\[-1mm]
\includegraphics[height=1.43in]{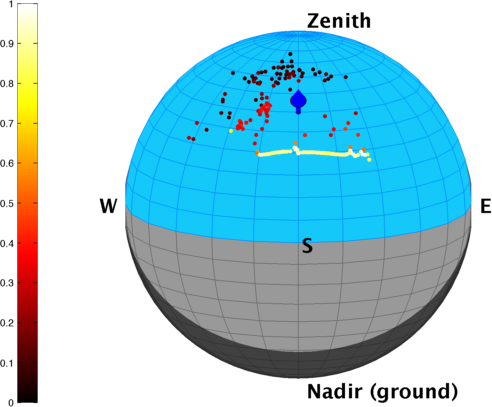} \ \ 
\includegraphics[height=1.43in]{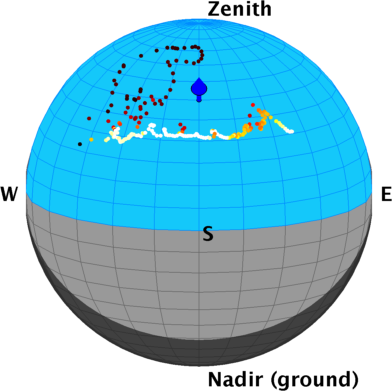} \\[3mm]
\includegraphics[height=1.43in]{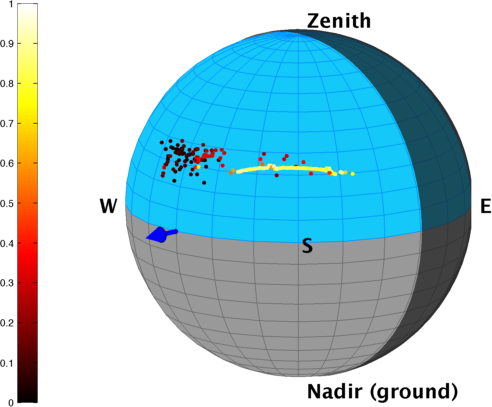} \ \ 
\includegraphics[height=1.43in]{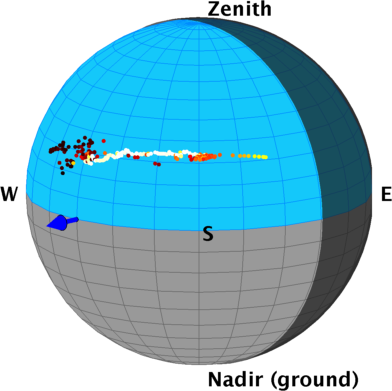} \\[3mm]
\includegraphics[height=1.43in]{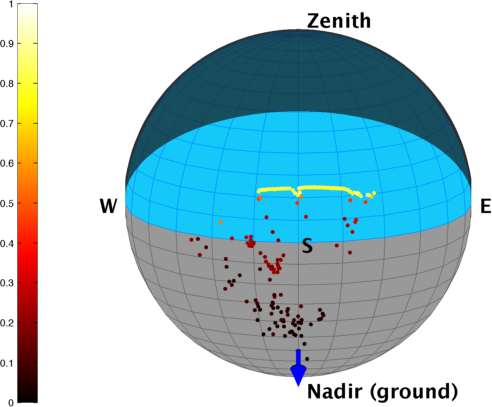} \ \ 
\includegraphics[height=1.43in]{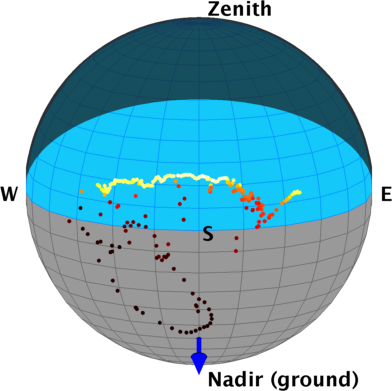} \\[2mm]
{\footnotesize {\verb| |} \hspace{0.5cm} (a) Mixed clouds (06-NOV-13) \hspace{0.25cm} (b) Mixed clouds (11-OCT-14) }\\
\caption[Examples of mean light vectors in function of the normal studied]{Globes representing the coordinate system of sky probes. Each normal (blue arrow) defines a shaded hemisphere in the environmental map that does not contribute light to the computed MLVs (dots). All MLVs in two particular partly cloudy days (columns) were computed from real environment maps~\cite{holdgeoffroy-iccp-15} for 3 example normal vectors (rows). Relative MLV intensities are shown in the color bar on the left.}
\label{fig:realShiftNormal}
\end{figure}

It is important to note that surface patches of different orientations (normals) are exposed to different hemispheres of illumination, with light arriving from above (sky) and below (ground). More MLV trajectories are shown in fig.~\ref{fig:realShiftNormal} for three different normal vectors (rows) and two different days (columns). Each globe represents the coordinate system for the environment maps captured in a day. For each combination normal-day, the time-trajectory of computed MLV directions (dots) and intensities (colors) are shown on the globe. Brighter MLVs lie close to the solar arc, while darker MLVs may shift away from it. Note that we present normals that are mainly Southward as they receive the most direct sunlight throughout the day in the Northern hemisphere. Surfaces with normals pointing North, for example, would be in shadow throughout the day in latitudes higher than the Tropic of Cancer around the winter solstice. Thus, the remainder of this paper considers a camera pointing North.

To more closely match the scenario considered above, we scale these real MLVs so that the brightest one over all days (\ie, for the most clear sky) has unit-length. From fig.~\ref{fig:realShiftNormal}, also note that some MLVs are shifted very far from the solar arc but, as indicated by the darker colors, their intensity is dimmed considerably by cloud coverage; little improvement in conditioning is obtained from these MLVs. 

\begin{figure}[!h]
\centering
\includegraphics[height=1.16in]{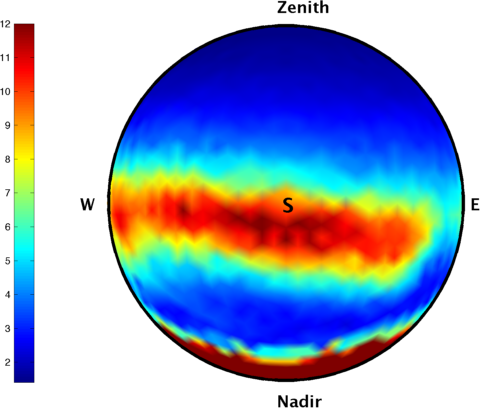} \ \ \ 
\includegraphics[height=1.16in]{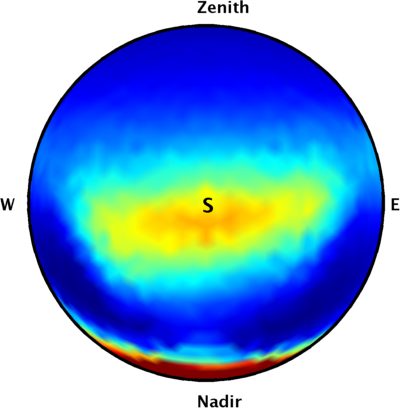} \\[1mm]
{\footnotesize {\verb| |} \hspace{0.2cm} (a) 06-NOV-13 \hspace{1.25cm} (b) 11-OCT-14 }\\[3mm]
\vspace{-1.5em}
\caption{Noise gain for normal directions ${\bf n}$ of patches visible to the camera, which is located South of the hypothetical target object. The colors indicate the shifting (coplanarity) of the associated MLVs. On both days, normals that are nearly horizontal are associated with nearly coplanar MLVs (smaller shifts, higher gains). These normals define a {\em zero-crossing region} between positive and negative out-of-plane shifts (mid row in fig.~\ref{fig:realShiftNormal}), where sun occlusion shifts MLVs predominantly {\em along} the solar arc.}
\label{fig:realShiftAll}
\end{figure}

Most important, fig.~\ref{fig:realShiftNormal} shows that the amount of out-of-plane MLV shift (elevation) relative to the solar arc also depends on the orientation ${\bf n}$ of the surface patch. This indicates that outdoor PS may present different degrees of uncertainty (conditioning) depending on the normal of each patch. Indeed, the maximum noise gain ($\lambda_{\max} = \max \left( \lambda_k \right)$) values in fig.~\ref{fig:realShiftAll} show that patches with nearly horizontal normals (orthogonal to the zenith direction) are associated with sets of MLVs that are closer to being coplanar throughout the day. As expected, patches oriented towards the bottom also present worse conditioning since they receive less light. 

%
%



This key observation also demonstrates the advantages of adopting more elaborate illumination models (\eg, \cite{yu-iccp-13}). For instance, the simpler point light model was used in \cite{shen-pg-14} to study the conditioning of outdoor PS. Because the atmospheric component is not modeled, the conclusion was that single-day reconstruction breaks down in two cases of nearly coplanar sun directions: closer to the poles near the winter solstice, and worldwide near an equinox. Our results suggest that more attention should be placed on the illumination model, without focusing exclusively on the sun.

\subsection{Lambertian, calibrated PS on partially cloudy days}
\label{sec:lambertian-ps}


The analysis performed on the HDR sky dataset (c.f. sec.~\ref{sec:overview}) indicates that surface patches may be better reconstructed in certain conditions, dependent upon cloud coverage and the orientation of the patch itself. In the case of partially cloudy days, our investigation reveals that those conditions usually shift the MLVs enough for outdoor PS methods to work.

\begin{figure*}[!ht]
 \centering
 \setlength{\tabcolsep}{0pt}
 \newcommand{\customwidth}{.072\linewidth}
 \begin{tabular}{@{}rcccccccccc@{}}
 &
 \begin{minipage}{\customwidth}\centering\scriptsize 10:30 \end{minipage} &
 \begin{minipage}{\customwidth}\centering\scriptsize 11:00 \end{minipage} &
 \begin{minipage}{\customwidth}\centering\scriptsize 11:30 \end{minipage} &
 \begin{minipage}{\customwidth}\centering\scriptsize 12:00 \end{minipage} &
 \begin{minipage}{\customwidth}\centering\scriptsize 13:00 \end{minipage} &
 \begin{minipage}{\customwidth}\centering\scriptsize 14:00 \end{minipage} &
 \begin{minipage}{\customwidth}\centering\scriptsize 14:30 \end{minipage} &
 \begin{minipage}{\customwidth}\centering\scriptsize 15:00 \end{minipage} &
 \begin{minipage}{\customwidth}\centering\scriptsize 15:30 \end{minipage} &
 \begin{minipage}{\customwidth}\centering\scriptsize 16:00 \end{minipage}
 \\
 \begin{sideways}\begin{minipage}{\customwidth}\centering \scriptsize illumination \vspace{5pt} \end{minipage}\end{sideways} &
 \includegraphics[width=\customwidth]{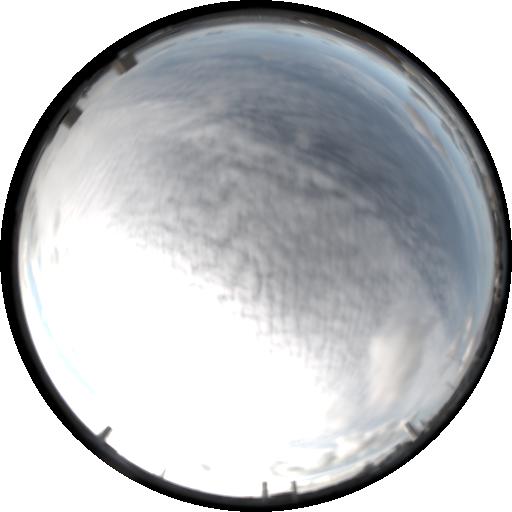} &
 \includegraphics[width=\customwidth]{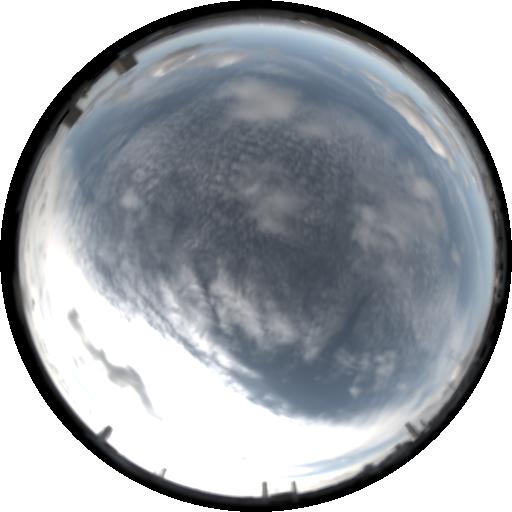} &
 \includegraphics[width=\customwidth]{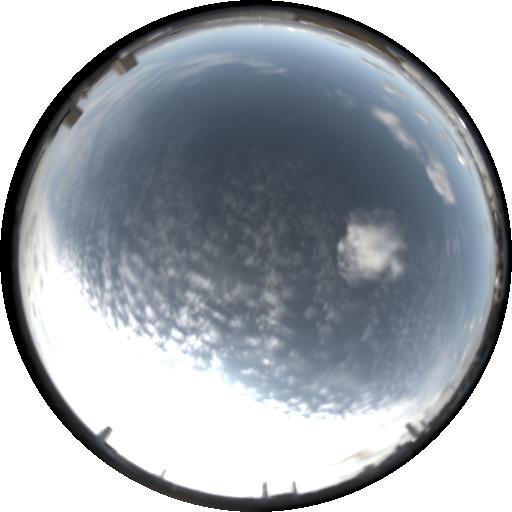} &
 \includegraphics[width=\customwidth]{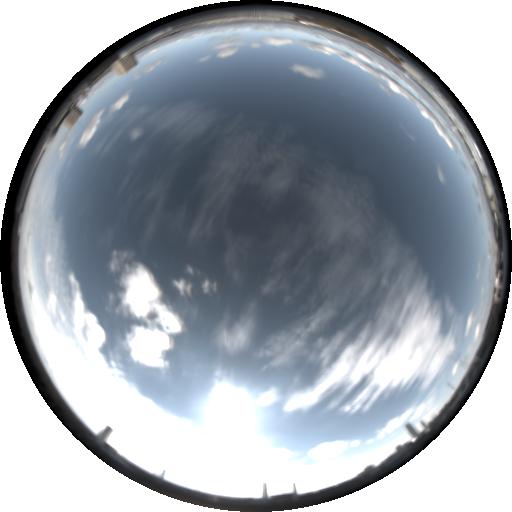} &
 \includegraphics[width=\customwidth]{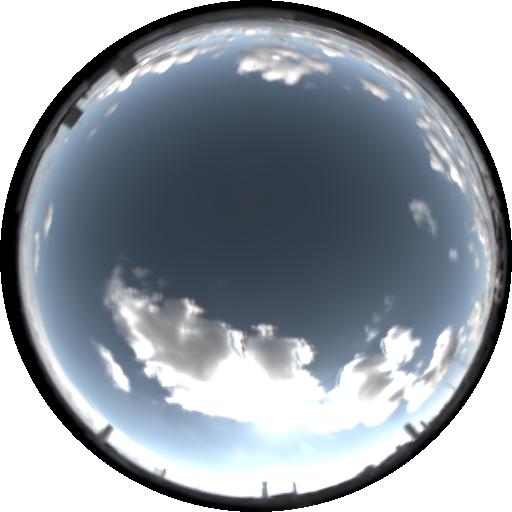} &
 \includegraphics[width=\customwidth]{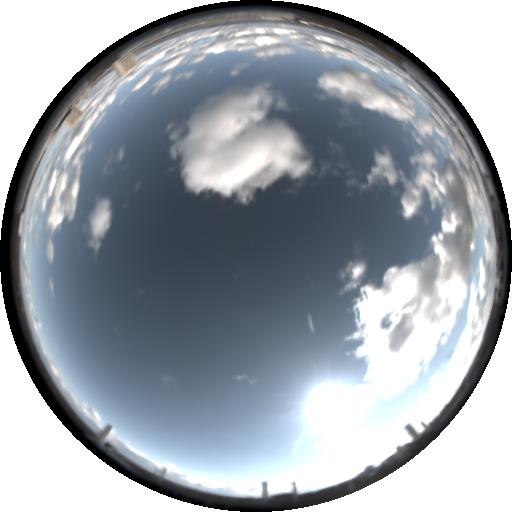} &
 \includegraphics[width=\customwidth]{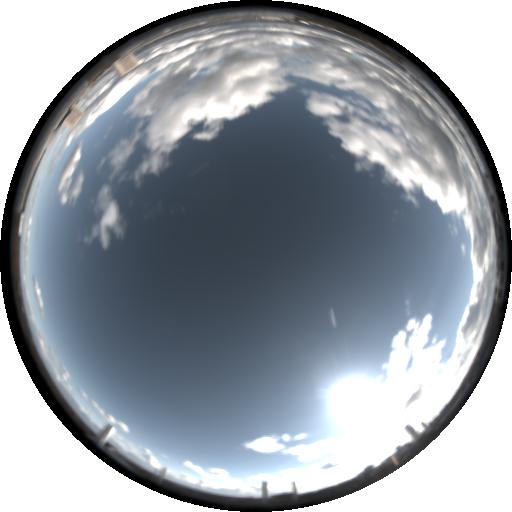} &
 \includegraphics[width=\customwidth]{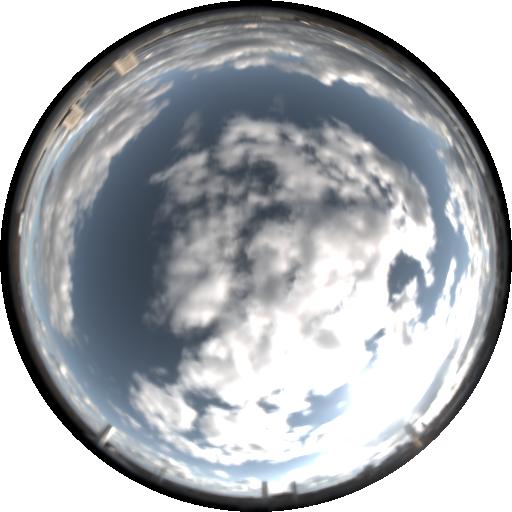} &
 \includegraphics[width=\customwidth]{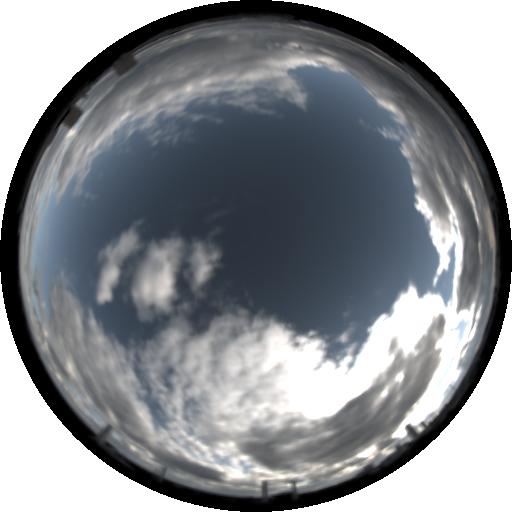} &
 \includegraphics[width=\customwidth]{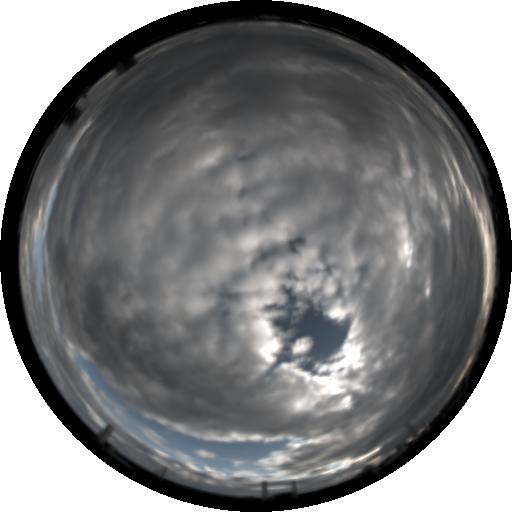}
 \\
 \begin{sideways}\begin{minipage}{.106\linewidth}\centering \scriptsize object \\ \vspace{1em} \vspace{5pt} \end{minipage}\end{sideways} &
 \includegraphics[width=\customwidth]{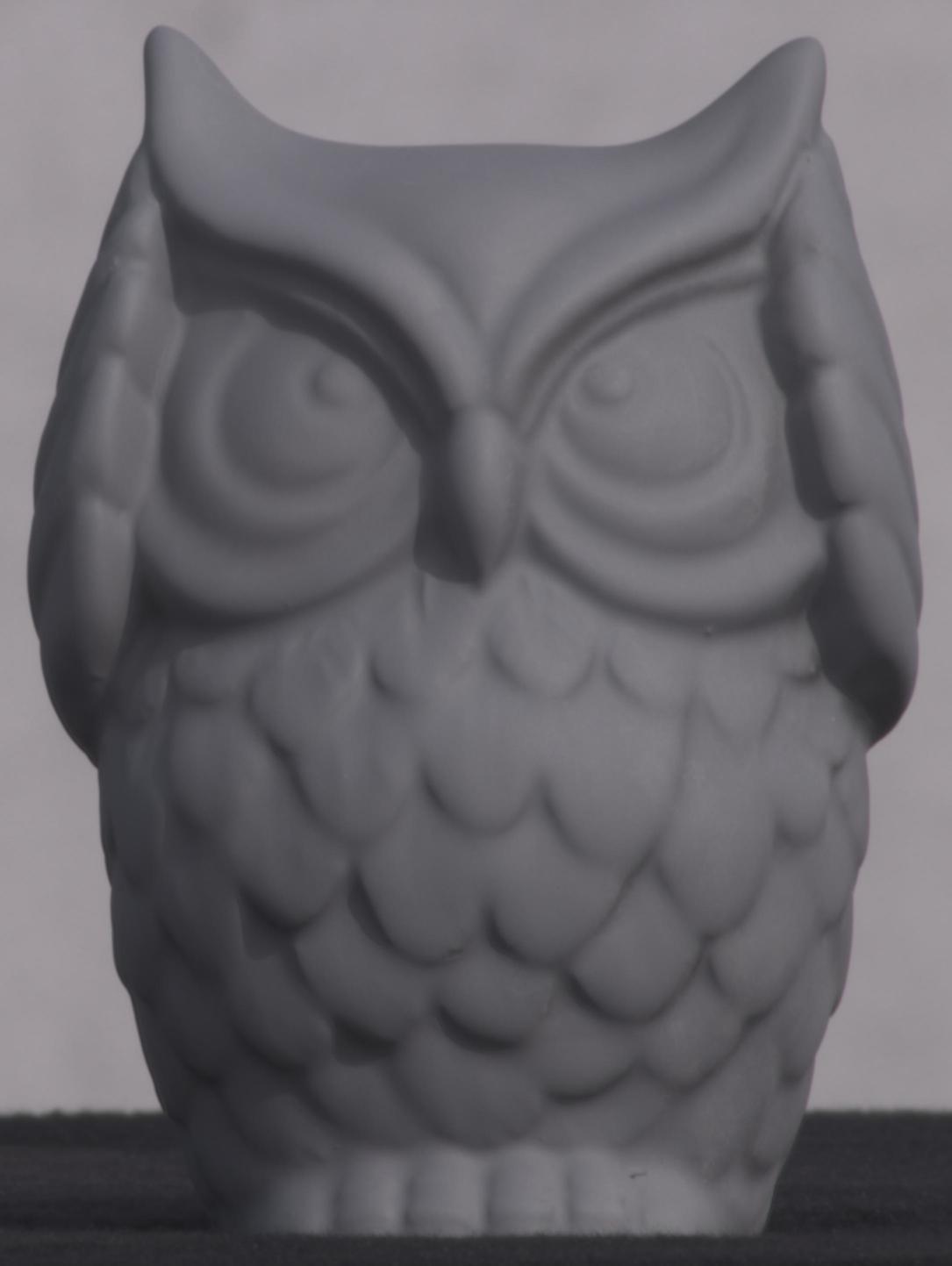} &
 \includegraphics[width=\customwidth]{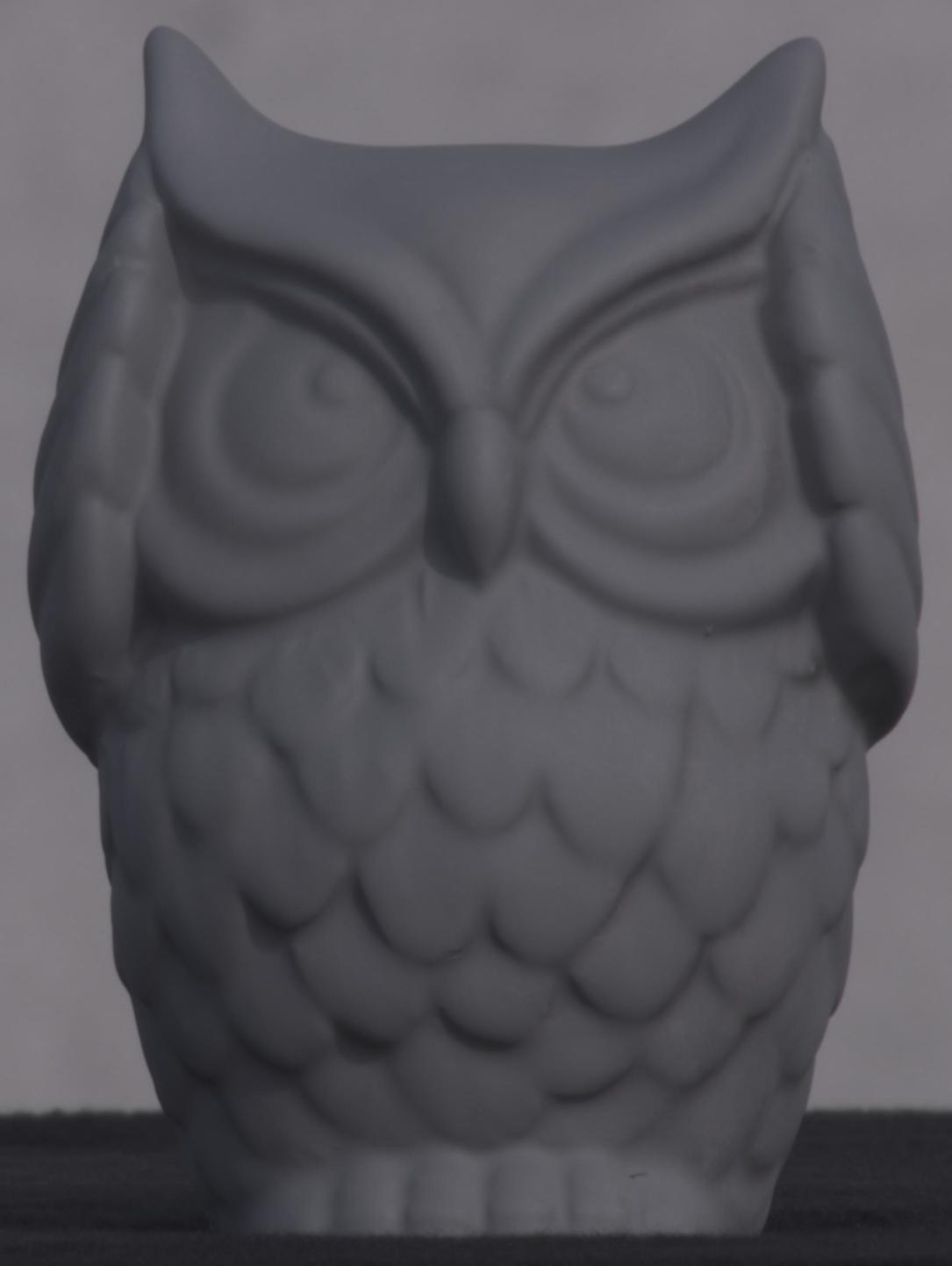} &
 \includegraphics[width=\customwidth]{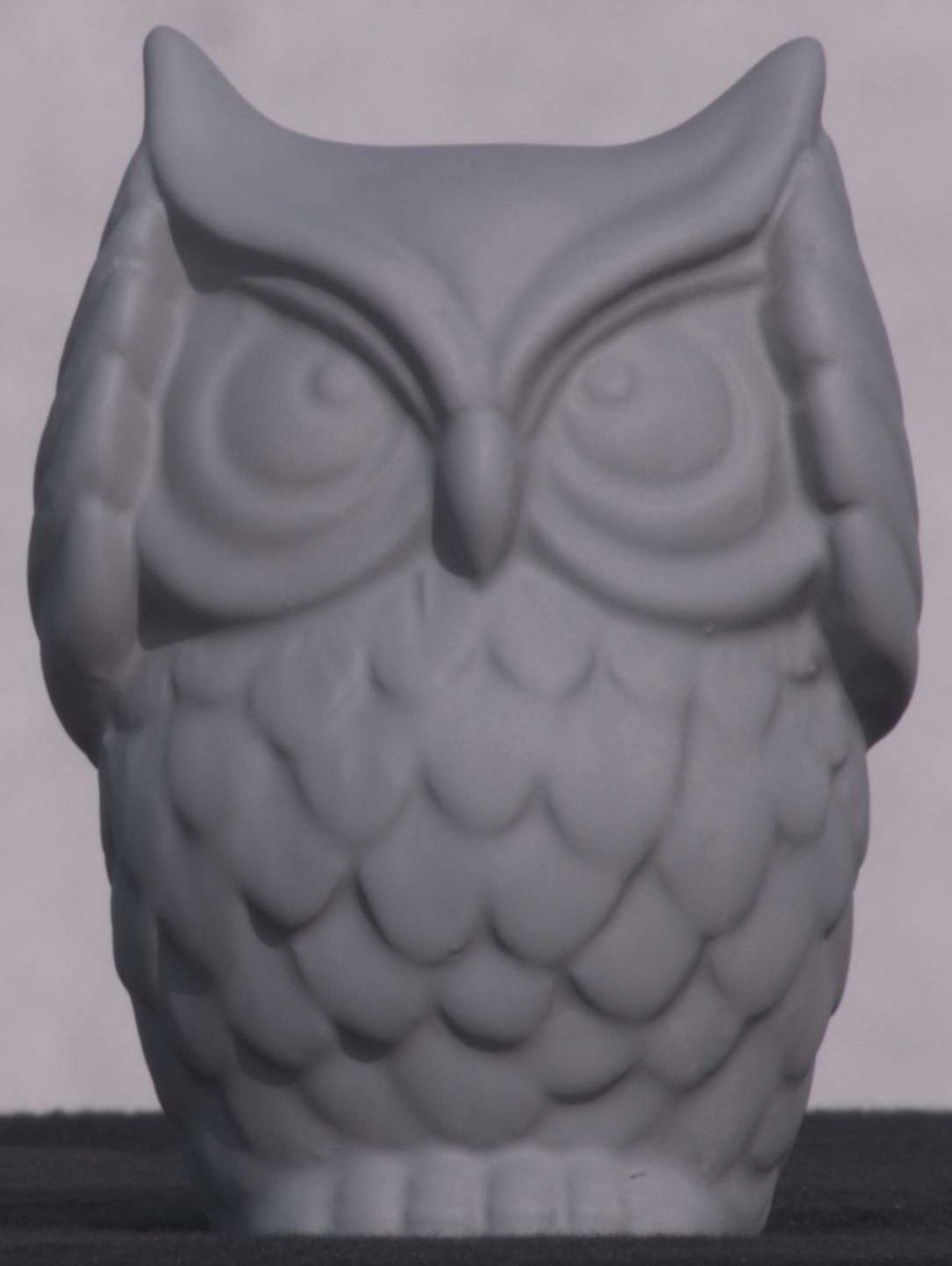} &
 \includegraphics[width=\customwidth]{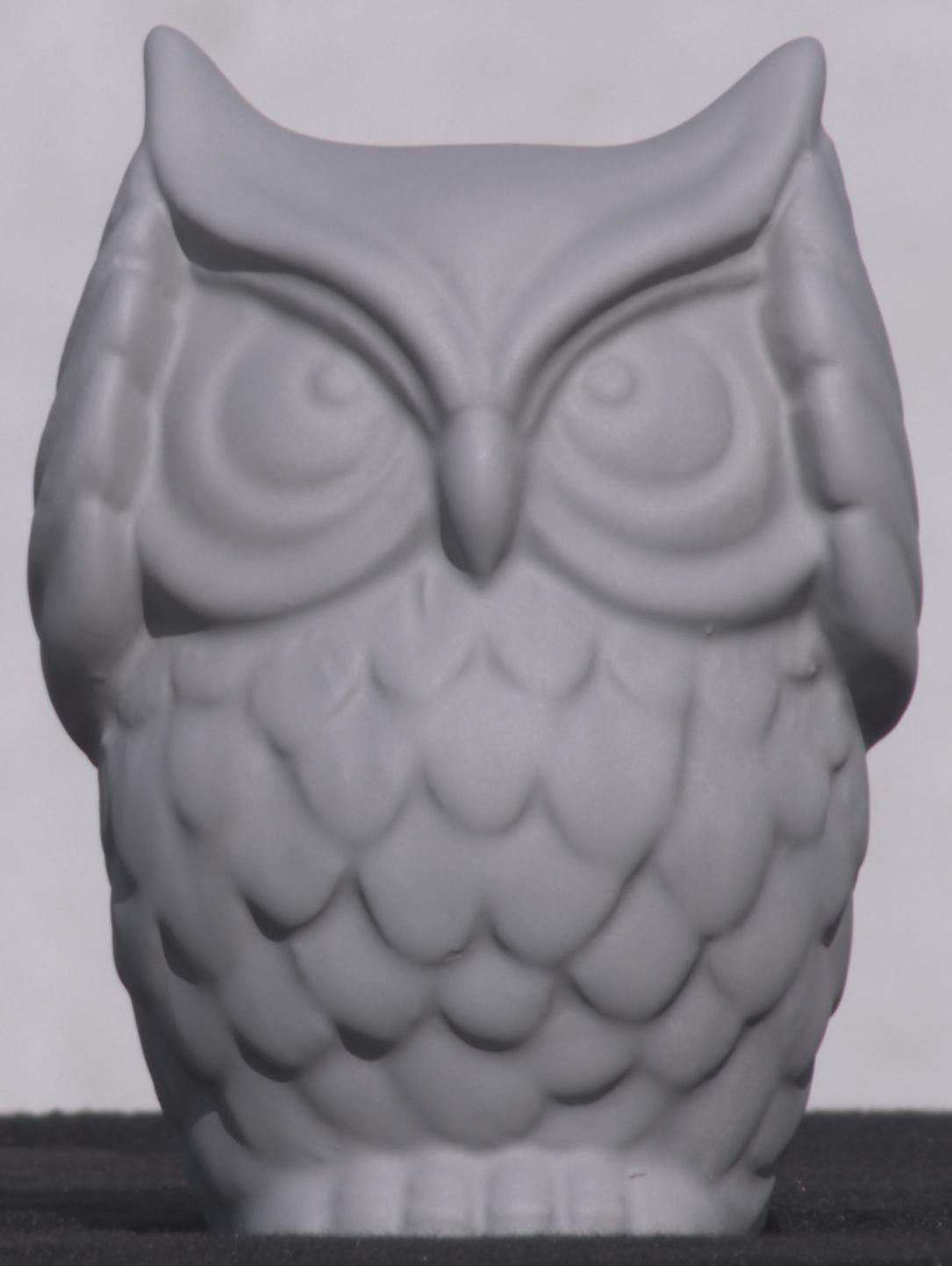} &
 \includegraphics[width=\customwidth]{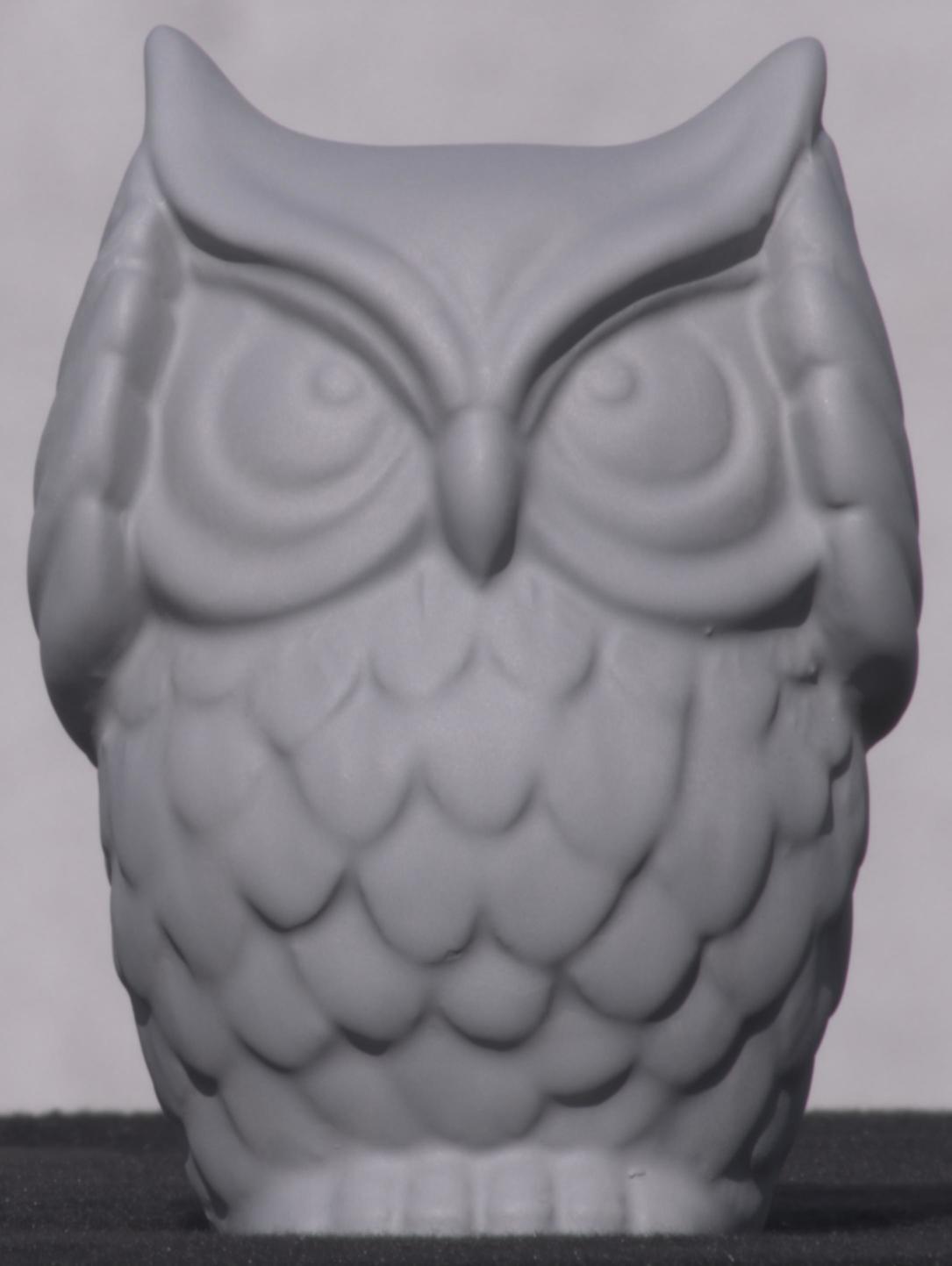} &
 \includegraphics[width=\customwidth]{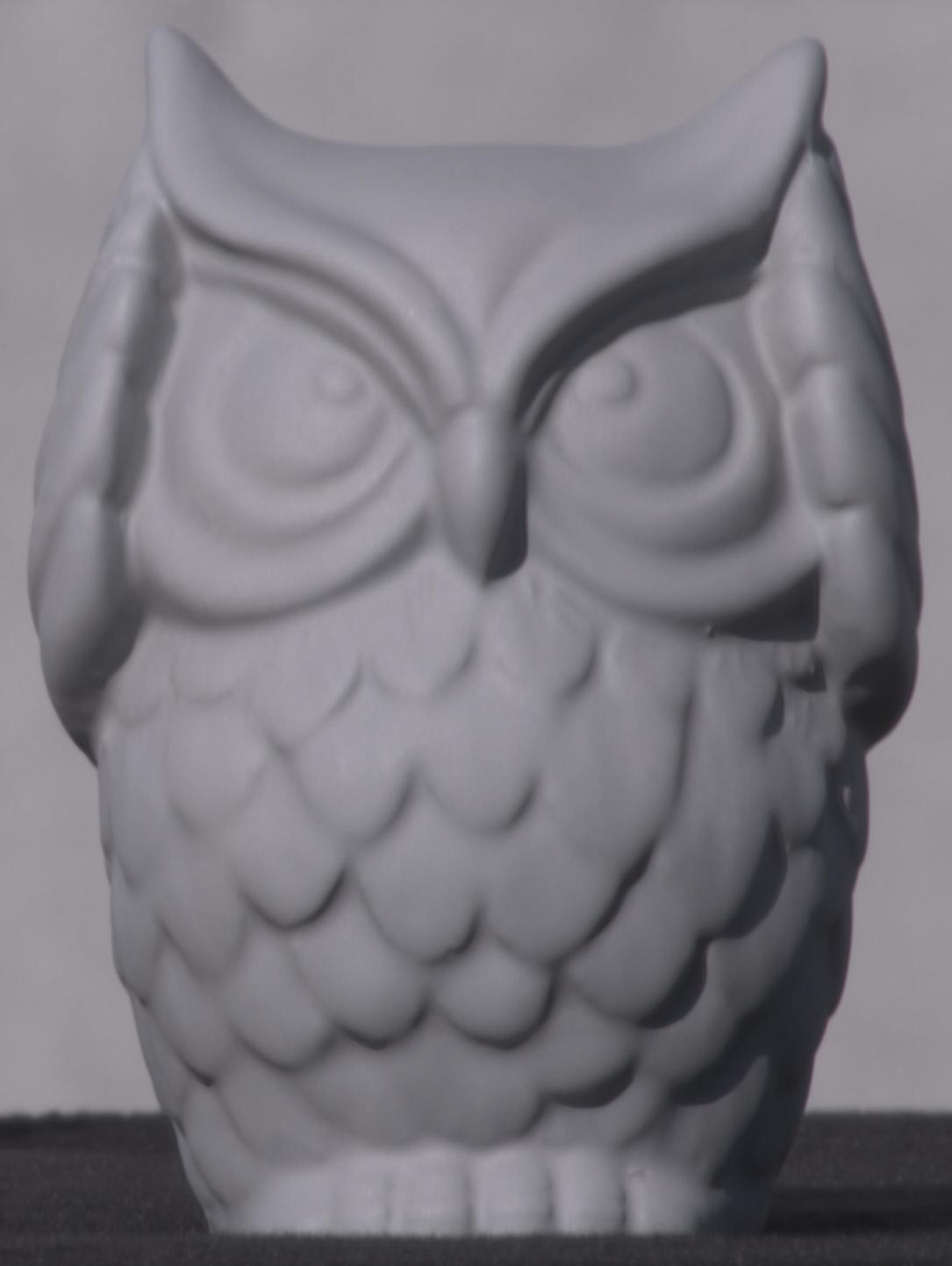} &
 \includegraphics[width=\customwidth]{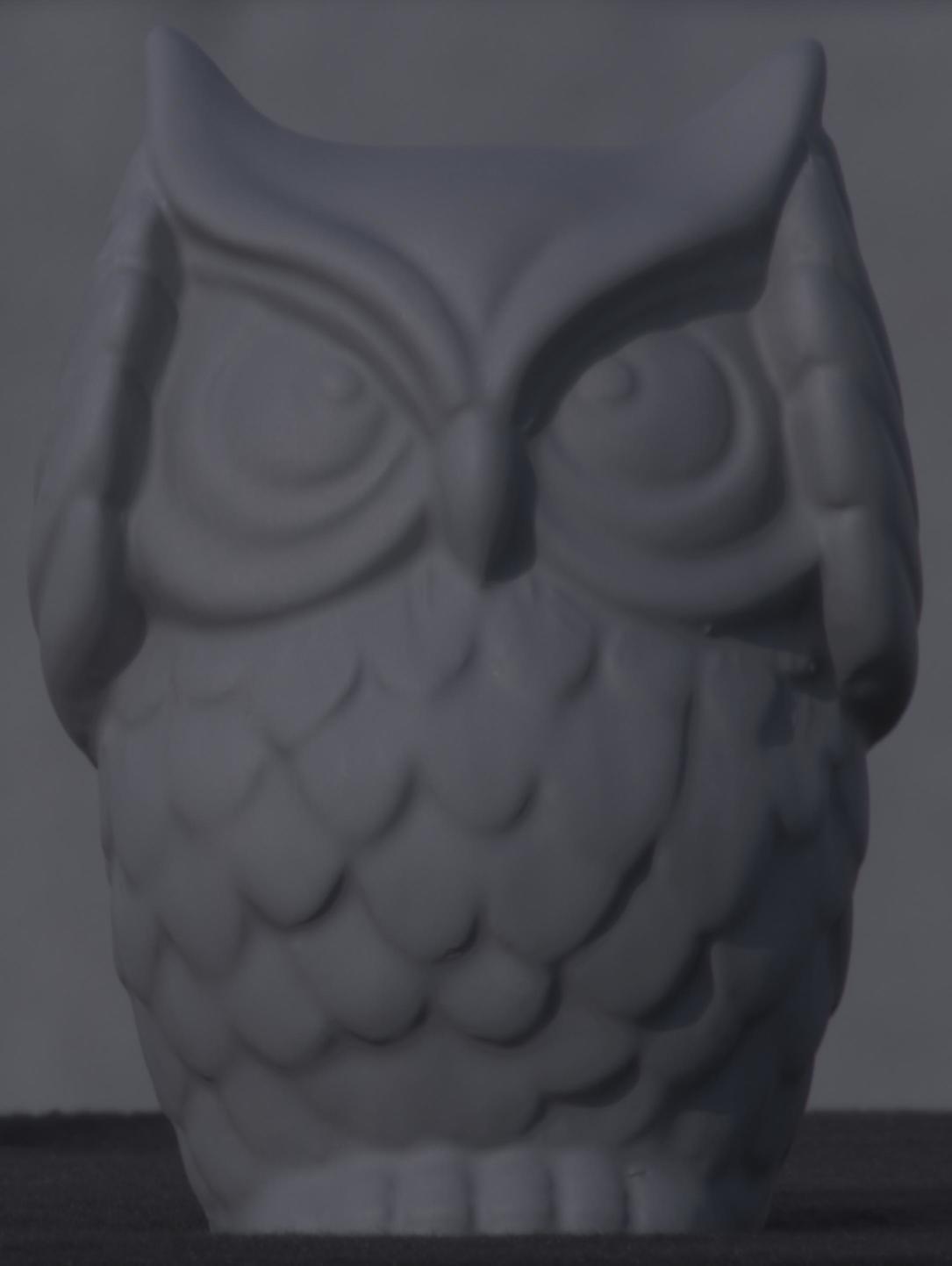} &
 \includegraphics[width=\customwidth]{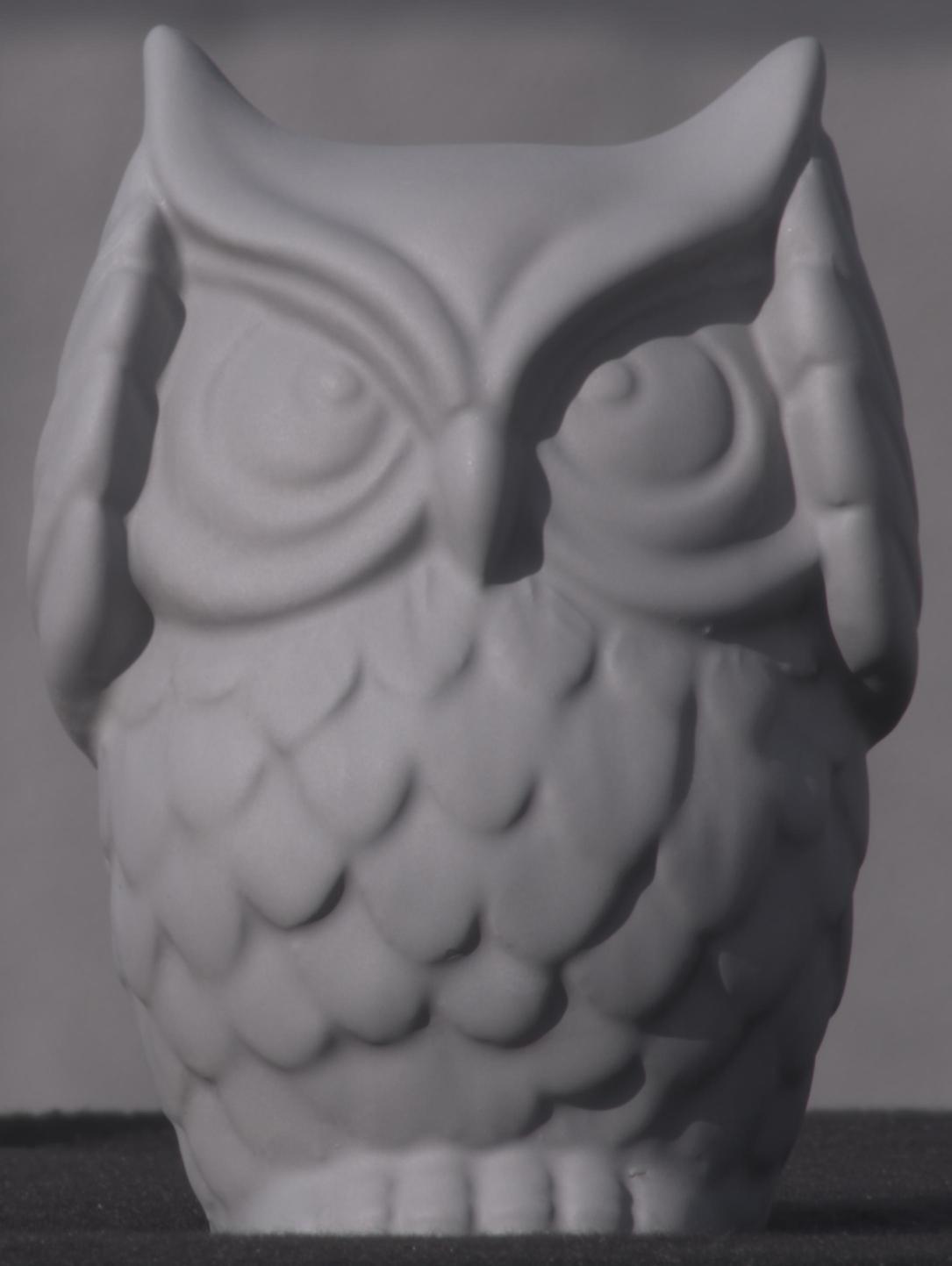} &
 \includegraphics[width=\customwidth]{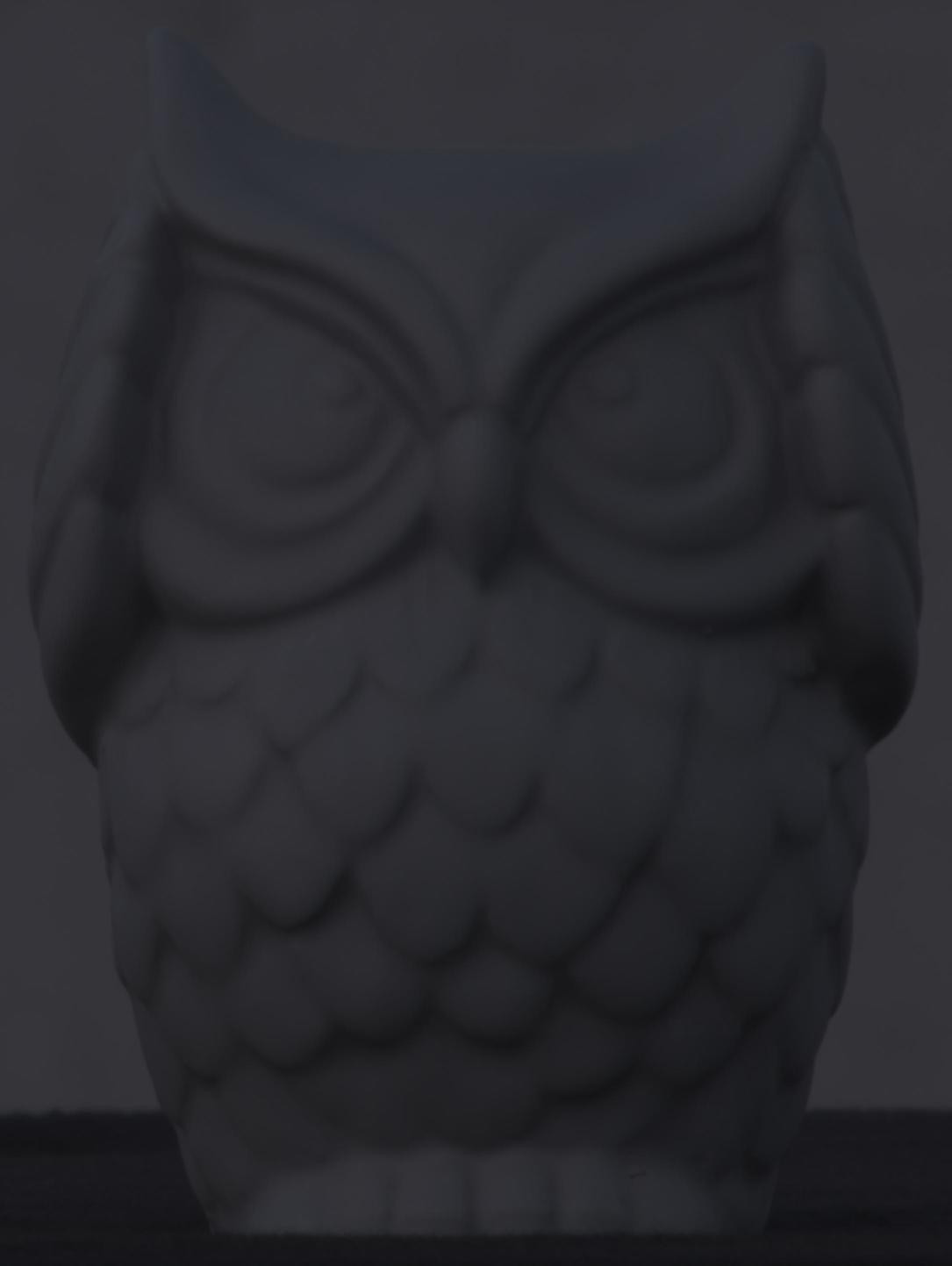} &
 \includegraphics[width=\customwidth]{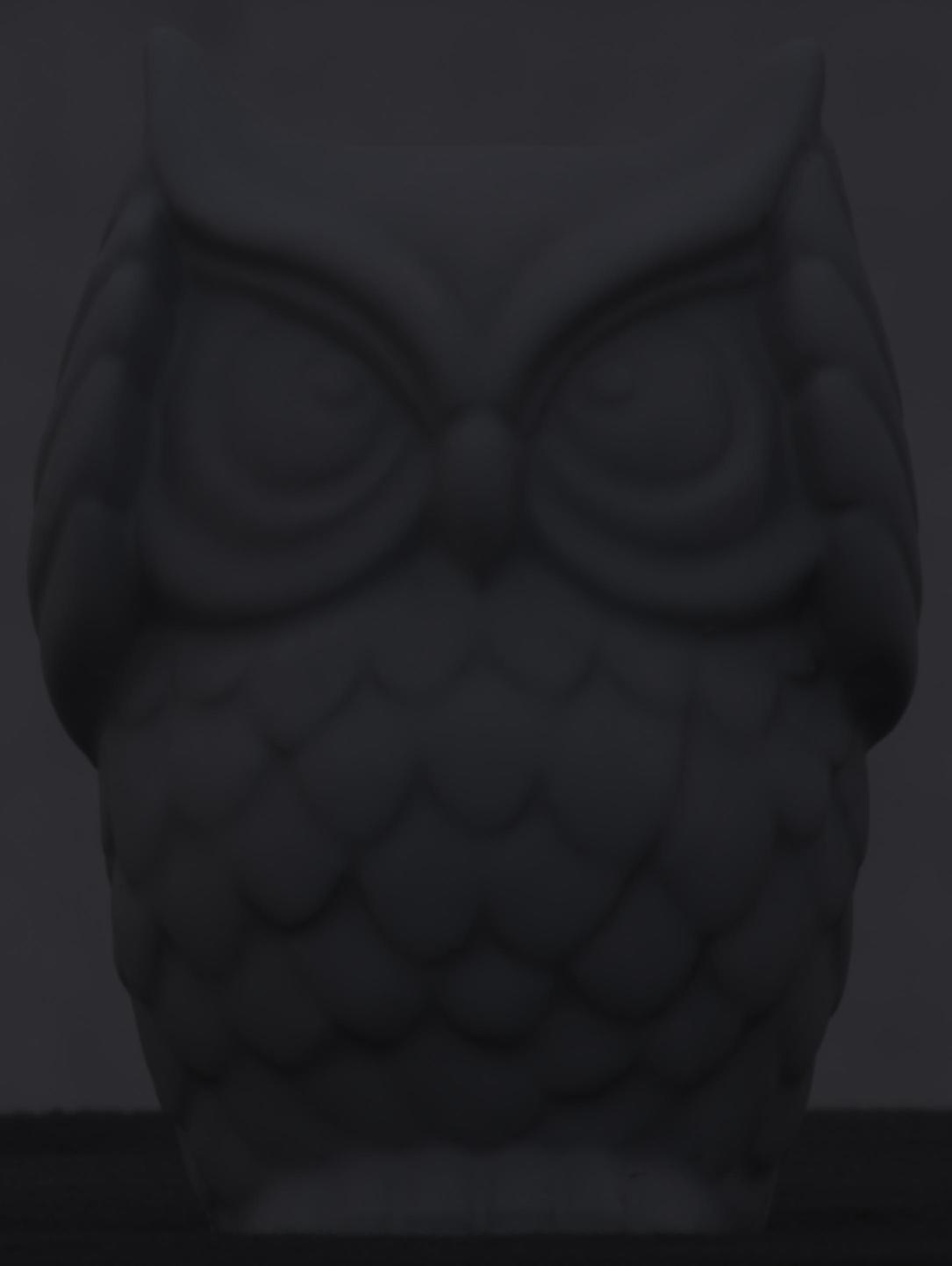}
 \end{tabular}
 \begin{minipage}{3.8cm}\centering\includegraphics[height=3.8cm]{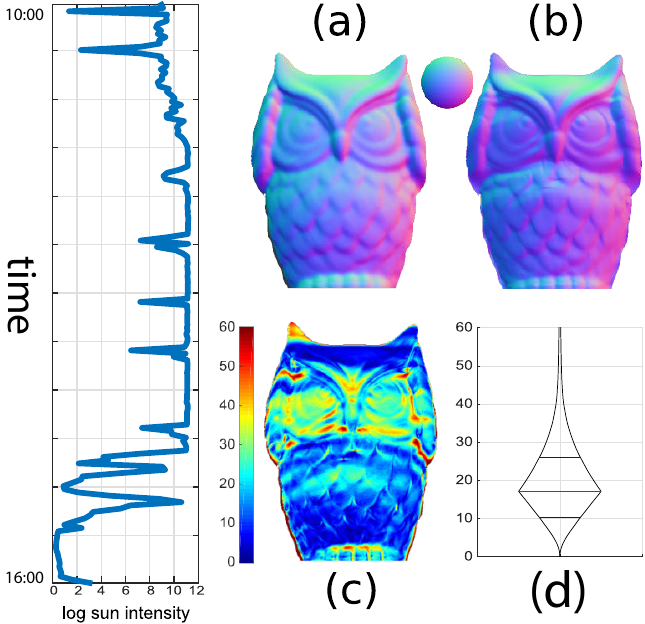}\end{minipage}
 
 \caption[Real data]{(left) Real outdoor HDR images of owl statuette and corresponding HDR environment maps (top row) providing synchronized, high-fidelity estimates of illumination conditions. All images were acquired on 10/11/2014 and tone-mapped for display only (with $\gamma = 1.6$). The sun visibility was 43\% on this day. We show the ground truth normals of the object (a) as well as normals recovered from~\cite{yu-iccp-13} (b), along with a reference normal sphere in inset. The reconstruction error (c) shows sphere is shown as a color coding reference; (b) normal estimation error at each pixel; and (c) the error distribution, in degrees.}
 \label{fig:reconstruction:example_envmaps}
\end{figure*}


To validate that accurate surface reconstructions can indeed be obtained on partially cloudy days, we captured a sequence of a real object lit by the sky over the course of a day. We oriented an owl statuette towards south and took 66 HDR captures using a Canon EOS Rebel SL1 between 10h30 and 16h30, local time, in Quebec City. We simultaneously captured hemispherical HDR sky images (as in sec.~\ref{sec:overview}) to provide high fidelity estimates of the illumination conditions for each image as shown in fig.~\ref{fig:reconstruction:example_envmaps}. Ground-truth surface normals were obtained by aligning a 3D model of the object (obtained with a Creaform MetraSCAN scanner) to the image using POSIT~\cite{dementhon-ijcv-95}.

We then perform calibrated outdoor PS on these images using the algorithm proposed by Yu et al.~\cite{yu-iccp-13}, with the following three differences: (\emph{i}) we use all possible pairs of images to compute ratios, instead of selecting a single denominator image; (\emph{ii}) we apply anisotropic regularization~\cite{hernandez-pami-11} to mitigate the impact of badly-conditioned pixels on the surface reconstruction; and (\emph{iii}) remove the low-rank matrix completion preprocessing, which, in our experiments, caused slightly degraded performance. 

The result on these real images is shown on fig.~\ref{fig:reconstruction:example_envmaps}. As predicted in the analysis from fig.~\ref{fig:realShiftAll}, normals on the head and the bottom of the abdomen, pointing respectively up or down, are mostly accurately estimated. As can be observed in~\ref{fig:reconstruction:example_envmaps}~(a), clouds sometimes occlude the sun, which improves the conditioning of the problem to yield an acceptable result. Without clouds, this day would have lead to an unstable formulation of the photometric stereo problem, as it is close to the fall equinox, which corresponds to the worst case scenario with coplanar sun directions~\cite{shen-pg-14}.

\section{Non-Lambertian PS on cloudless days}
\label{sec:NonLambertianSemiCalibrated}

Full environment maps taken at short intervals are needed to analyze the case of partly cloudy days, as one needs to capture the precise moment when the sun is occluded by clouds. However, in the case of clear days, the photometric cue is much weaker but the general appearance of the sky is more predictable and can be modeled by physically-based sky models (as in \cite{jung-ijcv-19}). 
In this section, we present a novel approach using deep learning to handle the ambiguities that arise in outdoor PS on a single day with a clear sky.

Our CNN-based approach compensates for the lack of photometric constraints by modeling prior knowledge on object geometry, material properties, as well as their local spatial correlation and interaction with natural outdoor lighting. In order to build such knowledge base, one needs a large number of images depicting various objects lit by outdoor lighting throughout the day, over different geographic locations and days over the year; finally, the surface normal map of each object is also required. Unfortunately, no such large-scale dataset currently exists, so a natural choice is to synthesize realistic data to train our network. Next, we present our problem formulation, CNN architecture, followed by the training procedure and data generation.



\subsection{Illumination model: the solar analemma}
\label{sec:analemma}

We follow a semi-calibrated PS approach that does not require known lighting environments~\cite{yu-iccp-13} nor complete camera geolocation data~\cite{jung-ijcv-19}. Our method only assumes that: (1) the object images are captured at approximate predefined times of the day, $t \in \{t_1, t_2, \ldots, t_T \}$; (2) the sun is unobstructed by clouds at these times; and (3) the camera is orthographic and faces approximately North (or South). In sec.~\ref{sec:analysis}, we analyze the robustness of our network with respect to departures from these ideal conditions. 

Together, these assumptions constrain the sun position to lie within an ``8-figure'' subspace at each time $t$, known as a solar {\em analemma}, whose shape also varies with geographical location (fig.~\ref{fig:solar-analemma}). For a given time $t$, the sun may be positioned at different locations depending upon the selected date and latitude, as prescribed by the analemma. The neural network is thus expected to adapt to this (constrained) variability in sun position and associated intensity. As shown in fig.~\ref{fig:solar-analemma}-(a,b), for a given timestamp and latitude, the sun position spans relatively small angular ranges, which still remain quite constrained even when considering geographical locations sampled over the Northern Hemisphere (fig.~\ref{fig:solar-analemma}-(c)) (note that a similar plot would be obtained by sampling the Southern Hemisphere with the camera facing South). 





Clear days can be accurately synthesized by parametric sky models, with much lower dimensionality in comparison to a full environment map. To generate training data, we use the physically-based parametric sky model of Ho\v{s}ek and Wilkie~\cite{hosek-tog-12} to obtain the spherical illumination function ${\bf L}_t(\mathbf{\boldomega})$ in eq.~\eqref{eqn:imageformation-continuous}. The model represents the spectral sky radiance as a parametric function of the sun position, sky turbidity and ground albedo; turbidity is set to 2, which corresponds to a clear day, and ground albedo to 0.3. Note that we do not model light scattering caused by clouds obscuring the sun and thus assume the sun is fully visible in the sky.

\subsection{Deep outdoor PS network}
\label{sec:proposed_method}




%
%

Here, we consider a more general image formation model in which the Lambertian term $\frac{\rho}{\pi}$ in eq.~\eqref{eqn:imageformation-continuous} is replaced with a standard GGX shader $\rho(\cdot)$ with varying diffuse and specular parameters. Our goal now is to invert this new rendering equation and recover the surface normal ${\bf n}$ based on the observed changes in pixel intensities $b_t$, which are caused by the changing natural illumination ${\bf L}_t(\mathbf{\boldomega})$ throughout the day. However, a solution based solely on the photometric cues from a sunny day is typically undefined due to limited sun motion and, thus, insufficient variability in ${\bf L}_t(\mathbf{\boldomega})$ and $b_t$.

Therefore, instead of considering a single pixel $b_t$, we reformulate our goal and instead aggregate additional RGB image data within a \emph{neighborhood} ${\bf B}_t \in \Reals^{P \times P \times 3}$, depicting a larger surface patch of width $P$ centered at the pixel $\mathbf{b}_t$. Now we seek to learn a predictor ${\bf N} = f({\bf B}_{t_1}, \ldots, {\bf B}_{t_T})$, where $T$ denotes the number of input images and ${\bf N} \in \Reals^{P\times P\times 3}$ is the patch normals. In this paper, $T=8$ and $P=16$  but we experiment with other values in secs.~\ref{sec:ablation_study} and \ref{sec:patch-size} respectively. This approach is motivated by the fact that complex object geometry is often made up of simpler, small surface patches presenting highly correlated surface normals and material properties.

A natural way to obtain this predictor $f(\cdot)$ is to train a Convolutional Neural Network (CNN) and learn a nonlinear function of local surface features that are highly correlated with the normal ${\bf n}$ at the center of the patch. We train our network on a large synthetic database of surface patches realistically rendered (sec.~\ref{sec:training_dataset}).




\begin{figure*}[t]
	\centering
	\includegraphics[width=1\linewidth]{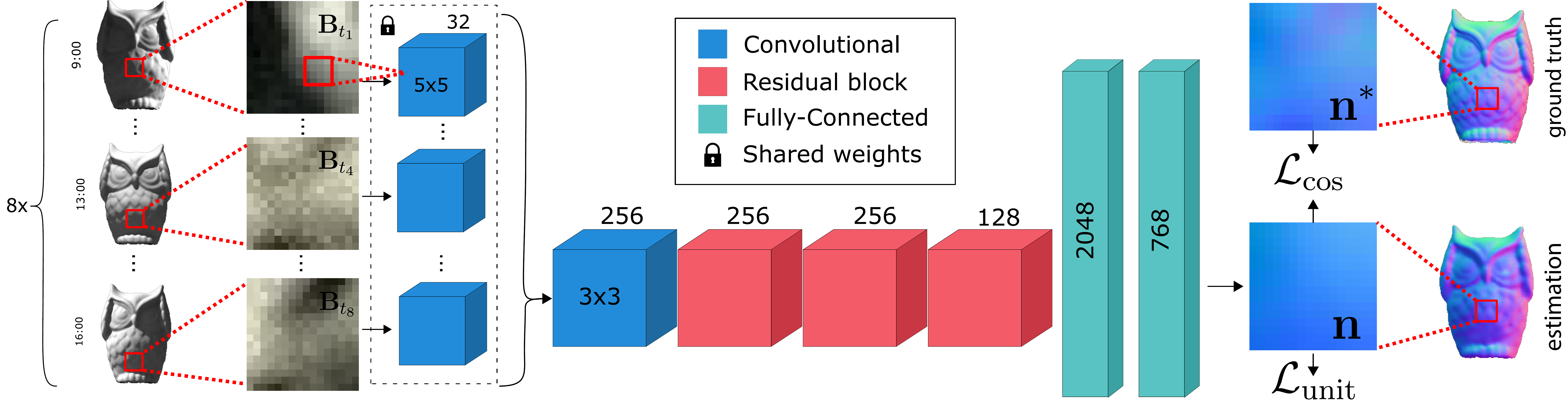}
	\vspace{-1em}
	\caption{Our novel CNN architecture for deep single-day outdoor PS on sunny days. The network operates on $16 \times 16$ patches $\mathbf{B}_t$ of the input image, captured at 8 time intervals $t$ regularly spaced throughout a single day. The network uses convolutional (blue) and residual (red) layers before estimating the normals using fully-connected layers (green). Two losses are used to train our method, one based on the cosine distance with the ground truth $\mathbf{\hat{n}}$ and another to constrain the norm of the output vector.}
	\label{fig:architecture}
\end{figure*}

\subsubsection{Network architecture}
\label{sec:architecture}

The function ${\bf N} = f({\bf B}_{t_1}, \ldots, {\bf B}_{t_T})$ introduced above is designed as CNN with the architecture shown in fig.~\ref{fig:architecture}. The network takes $8$ input $16 \times 16$ image patches, extracted from 8 images captured at regular intervals $\Delta t$ between 9:00 and 16:00 solar time throughout a single sunny day. Note that no other information (geolocation, capture time, etc.) is provided to the network. The first layer is composed of 32 channels of $5 \times 5$ filters with shared weights across the 8 inputs. The resulting feature maps are subsequently concatenated in a single $14 \times 14 \times 256$ feature tensor. A second convolutional layer is then used, yielding 256 channels, followed by 3 residual blocks as defined in the resnet-18 architecture~\cite{he-cvpr-16}. Lastly, 2 fully-connected layers (FC) are used to produce a $16 \times 16 \times 3$ patch of estimated normals $\mathbf{n}$. Note that we experimented with fully-convolutional architectures~\cite{taniai-arxiv-18} but found the FC layers to yield better results. The ELU activation function~\cite{clevert-iclr-16} is used at every convolutional and fully connected layer, except the output layer where a $\tanh(\cdot)$ function is used. As in~\cite{isola-cvpr-2017}, batch normalization~\cite{ioffe-icml-15} is applied at every layer except the first and the output layer.



The $16 \times 16$ estimated normals are represented by Cartesian $(x,y,z)$ components of the surface normal of the input patch. We experimented with parameterization in both Cartesian $(x,y,z)$ and spherical coordinates $(\theta,\phi)$, but found the former to be more stable despite its additional degree of freedom. We hypothesize this may be due to the ``wrap-around'' issue with the azimuth angle $\phi$. 

To process entire images, we crop overlapping tiles from the image with a stride of 8 pixels. Since a pixel can belong to up to 4 patches, the network produces several estimates $\mathbf{\hat{n}}$ that are then merged together using a weighted average. We use a Gaussian kernel with $\sigma=4$ centered on the middle of the patch as weighting function to perform the linear interpolation across overlapping patches. 

\subsubsection{Training}

The network learns a function that estimates the patch normals $\mathbf{N}$. We define the loss to be minimized between the estimated and ground truth patch normals $\mathbf{N}$ and $\mathbf{N}^*$ respectively as the sum of two separate loss functions defined on individual patch normals $\mathbf{n}_i$, $i \in \{1, ..., N\}$ where $N = 16 \times 16 = 256$. The total loss is the sum over all $N$ individual normals: 
\begin{equation}
\mathcal{L}(\mathbf{N}, \mathbf{N}^*) = \sum_{i=1}^{N} \left(\mathcal{L}_{\cos}(\mathbf{n}_i, \mathbf{n}^*_i) + \mathcal{L}_{\mathrm{unit}}(\mathbf{n}_i) \right)\,.
\label{eqn:loss}
\end{equation}
The first term is the cosine distance between the estimated $\mathbf{n}_i$ and ground truth normal $\mathbf{n}^*_i$:
\begin{equation}
\mathcal{L}_{\cos}(\mathbf{n}_i, \mathbf{n}^*_i) = 1 - \frac{ \left\langle \mathbf{n}_i , \mathbf{n}^*_i \right\rangle }{ \lVert \mathbf{n}_i \rVert \lVert \mathbf{n}^*_i \rVert } \,,
\end{equation}
where $\left\langle \cdot , \cdot \right\rangle$ denotes the dot product. The second term enforces the unit-length constraint on the recovered normal: 
\begin{equation}
\mathcal{L}_{\mathrm{unit}}(\mathbf{n}_i) = \left| \; \lVert \mathbf{n}_i \rVert - 1 \; \right| \,.
\end{equation}
The loss in eq.~\eqref{eqn:loss} is minimized via stochastic gradient descent using the Adam optimizer~\cite{kingma-iclr-15} with an initial learning rage of $\eta = 0.001$, a weight decay $\lambda = 1\times10^{-4}$ and the recommended values $\beta_1 = 0.9$ and $\beta_2 = 0.999$. Mini-batches of 128 samples were used during training and regularized via early stopping. Training typically converges in around 250 epochs on our dataset, which is described next.

\begin{figure}[t]
    \centering
    \includegraphics[width=\linewidth]{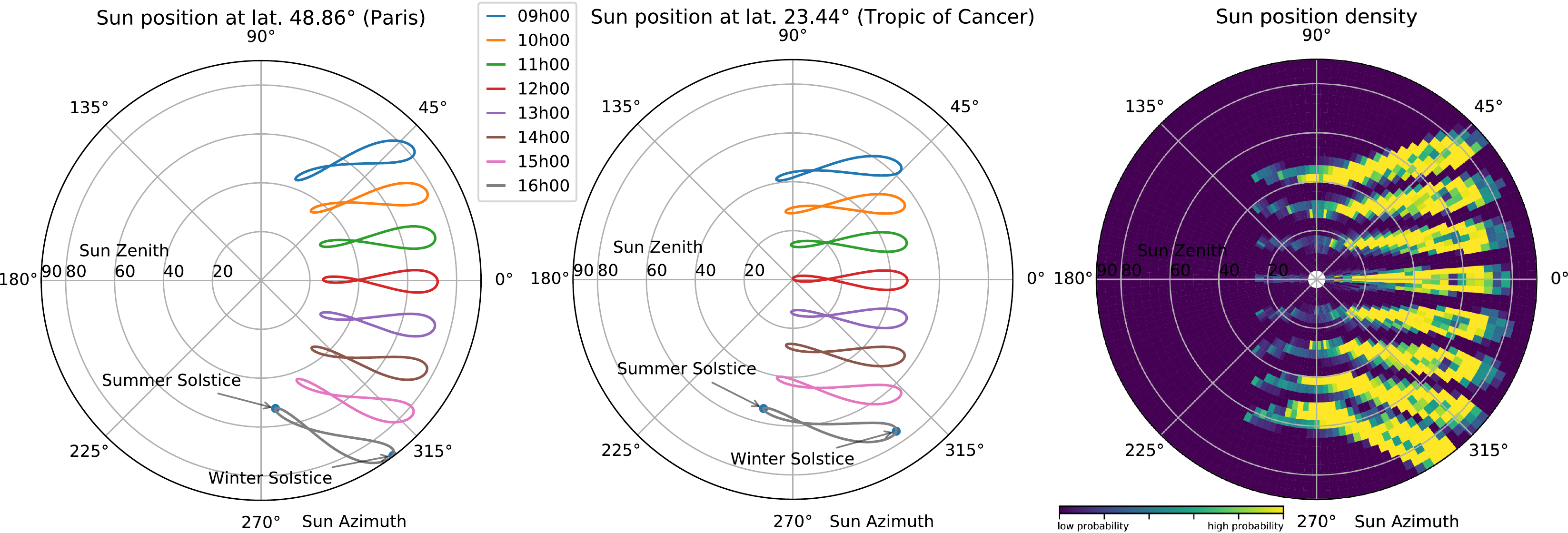}
    \begin{tabular*}{\linewidth}{c@{\extracolsep{\fill}}c@{}c}
        \hspace{3.2em}(a) & \hspace{0.0em} (b) & (c)\hspace{3.1em}
    \end{tabular*} \\ \vspace{-0.5em}
    \caption{Solar analemma: position of the sun in the sky at a specific time of the day and throughout a year over (a) Paris and (b) the Tropic of Cancer. Note how the analemmas spread over a wide range of zenith and azimuth angles over the course of a year. (c)~Probability of the sun location in the sky for our training set.}
    \label{fig:solar-analemma}
\end{figure}


\subsection{Training dataset}
\label{sec:training_dataset}

To train our predictor function $f(\cdot)$, we rely on a large training dataset of synthetic objects, lit by a physically-based outdoor daylight model. To generate a single 8-images set of inputs, we randomly select a combination of: 1) object shape, 2) material, and 3) geo-temporal coordinates for lighting. We now detail how each of these 3 choices are made. 

Since the neural network only sees patches of $16 \times 16$ pixels, its receptive field is, by design, not large enough to learn priors on whole object shapes. Therefore, our dataset contains a wide variety of local surface patches. We used the blob dataset from~\cite{johnson-cvpr-11} as training models. We also added simple  primitives (cube, sphere, icosahedron, cone) to the data. A validation set, comprised of one of the blobs models that was kept from the training set as well as some models from the Stanford 3D Scanning Repository~\cite{curless-cg-96} and the owl model used in~\cite{holdgeoffroy-iccp-15}, was also created. All blobs and geometric primitives are randomly rotated about their centroid. 


To model a wide range of surface appearance ranging from diffuse to glossy, we employ a linear combination of a lambertian and a microfacet model: 
\begin{equation}
\rho(\boldomega, {\bf v}, {\bf n}) = \boldrho_c (\alpha + (1-\alpha)\rho_\text{GGX}(\boldomega, {\bf v}, {\bf n}, \sigma)) \,,
\label{eq:brdf}
\end{equation}
where $\boldrho_c \in \Reals^3$ is the surface color, and $\rho_\text{GGX}$ is the GGX microfacet model~\cite{walter-eg-07} with surface roughness parameter $\sigma$. 

The albedo $\boldrho$ is generated in HSV space, where $H \sim U(0,1)$, $S \sim T(0, 0, 1)$, and $V \sim T(0, 0.75, 1)$, where $U(a, b)$ is a uniform distribution in the $[a, b]$ interval and $T(a,b,c)$ is a triangular distribution in the $[a,c]$ interval with mode $b$. This generates colors that are in general bright and prevents an abundance of strongly saturated colors. Surface roughness $\sigma$ is sampled as $\sigma \sim T(0.2, 0.4, 1)$ to avoid mirror-like surfaces. Finally, we sample a mixing coefficient $\alpha \sim U(0,1)$. 

To light the scene with a wide variety of realistic outdoor lighting conditions, we rely on the Ho\v{s}ek-Wilkie physically-based sky model~\cite{hosek-tog-12}. We also placed a ground plane of albedo 0.3 outside the field of view of the camera, to generate a light bounce from below the object. 11 random locations in the Northern Hemisphere between latitude $0\degr$ (Equator) and $56\degr$ (Moscow) were selected. Furthermore, 6 random days throughout the year were chosen in addition to the equinoxes and solstices. This results in 110 pairs of geographical locations and dates, which are used to compute the sun position in the sky throughout the day using \cite{bretagnon-aaa-88}. The distribution of the resulting sun positions throughout our training set is shown in fig.~\ref{fig:solar-analemma}. For every pair of geographical location and day, 8 timestamps ranging from 9:00 to 16:00 are used to perform the renders. Timestamps are aligned to the solar noon instead of the political time zone of the geographic location. Although we sample only geographical locations in the Northern hemisphere, our dataset represents equally well days in the Southern hemisphere. Indeed, flipping the images left-right, reversing the image order (from 16:00 to 9:00) and pointing the camera Southward would generate data identical to our training dataset.

The resulting images are rendered with the Cycles physically-based rendering engine. This results in a dataset of 369,440 renders corresponding to 23,090 combinations of geo-temporal coordinates and materials properties, which we then split into 21,220 and 1870 for training and validation, respectively. Each render has a resolution of $256 \times 256$ pixels, which amounts to over 10 millions input-output pairs of $16 \times 16$ patches to train on. Special care was taken into ensuring no 3D model nor material properties were shared between both the training and validation datasets. \textbf{Please see the supplementary material for example training images.}

\begin{figure}
\centering
\includegraphics[width=\linewidth]{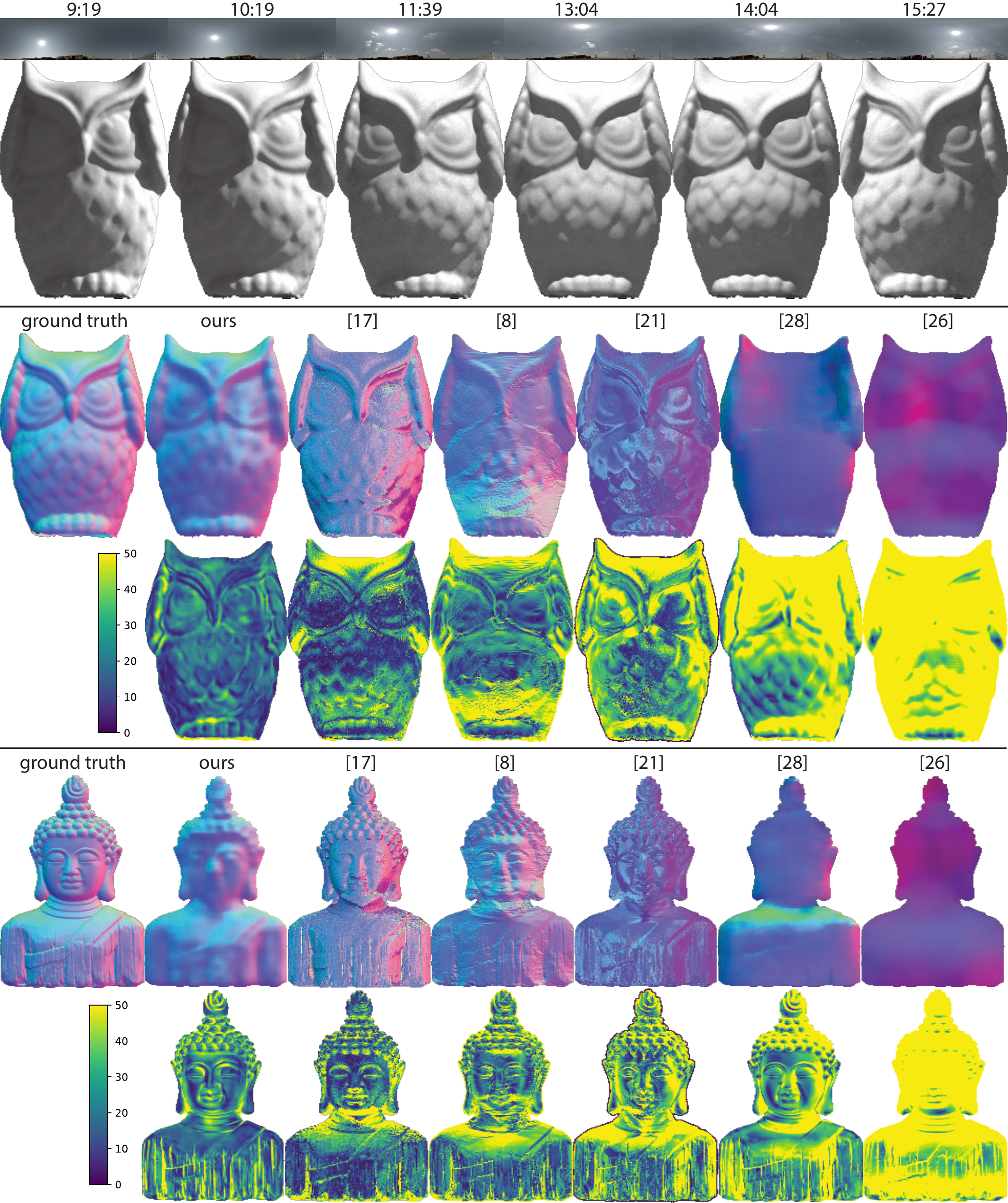}
\caption{(top) An example of the lighting environment maps and renders throughout a day. (bottom) Qualitative results (odd rows) and errors in degrees (even rows) of our technique and the state-of-the-art on single-day photometric stereo in the semi-calibrated~\cite{jung-ijcv-19} and calibrated~\cite{yu-iccp-13} cases, deep photometric stereo~\cite{santo-iccv-17} and single image normal estimation~\cite{wu-nips-17,eigen-iccv-15} (averaged over the day) on our real lighting dataset. \textbf{More results available in the supplementary material.}}
\label{fig:results-qualitative}
\end{figure}





\subsection{Results on synthetic images with real lighting}
\label{sec:evaluation_dataset}

We first evaluate and compare the techniques using a dataset of synthetic objects lit by real skies. To generate the images, we manually selected 3 sunny days over 2 geographical locations from the Laval HDR sky database~\cite{hdrdb}, which contains unsaturated HDR, omnidirectional photographs of the sky captured with the approach proposed in \cite{stumpfel-afrigraph-04}. We build a virtual 3D scene containing the HDR sky environment map as the sole light source, a 3D object viewed by an orthographic camera, and a 0.3 albedo ground plane placed under the object, outside the field of view of the camera. We used the 3D models from the validation set which the neural network never saw during training. This results in a dataset of 960 renders yielding 60 normal maps to evaluate. Example images obtained with this technique are shown in fig.~\ref{fig:results-qualitative}. 

\begin{figure}[t]
\centering
\includegraphics[width=.99\linewidth]{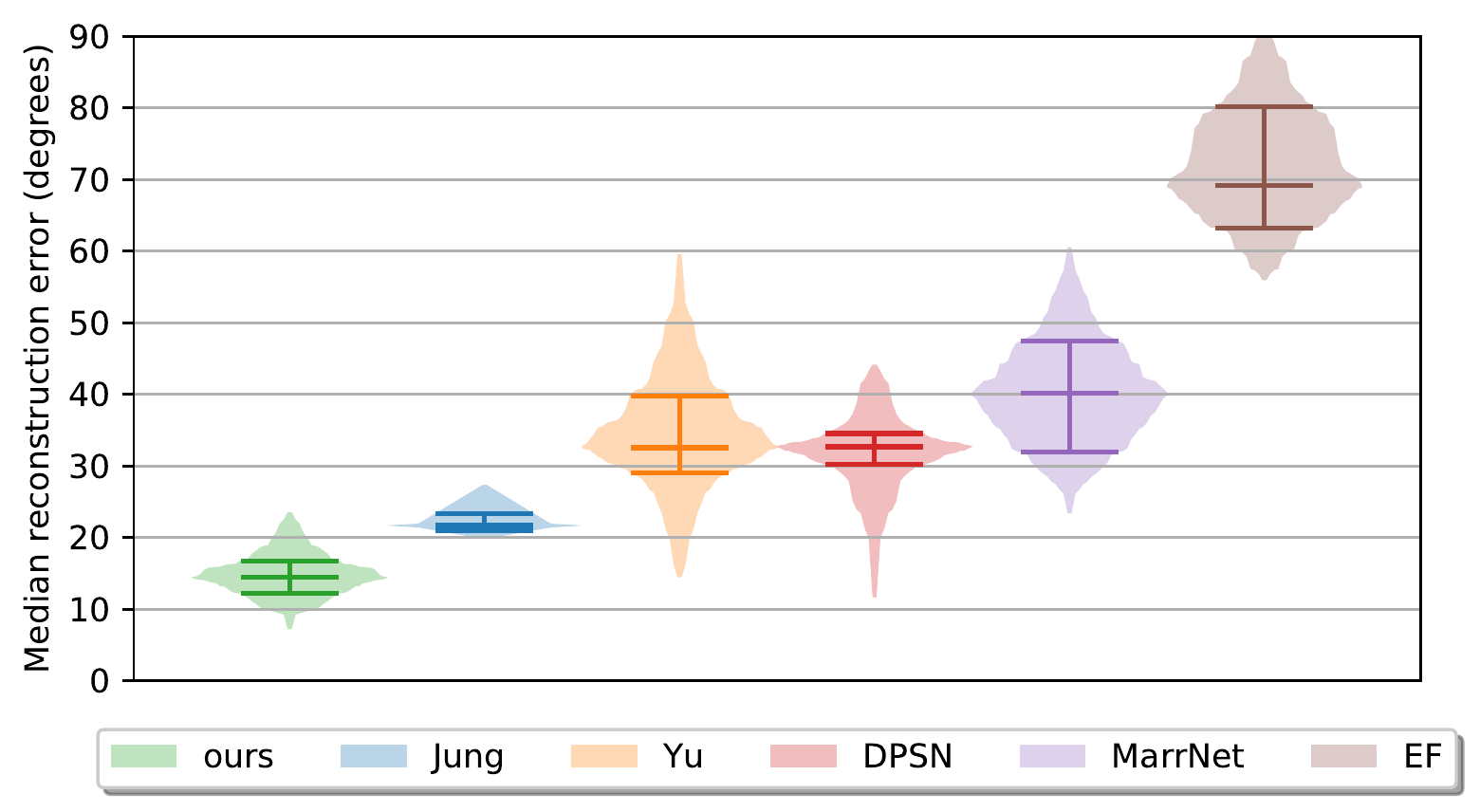}
\caption{Median reconstruction error on our real lighting dataset displayed vertically as “box-percentile plots”~\cite{esty-jss-03}; the center horizontal bars indicate the median, while the bottom (top) bars are the 25th (75th) percentiles. Our proposed method (green) provides state-of-the-art performance compared to non-learned methods for single-day PS (blue~\cite{jung-ijcv-19}, orange~\cite{yu-iccp-13}), deep learning methods on calibrated photometric stereo (red~\cite{santo-iccv-17}) and single image normals reconstruction (purple~\cite{wu-nips-17}, brown~\cite{eigen-iccv-15}).}
\label{fig:results-quantitative}
\end{figure}

We compare our method to several state-of-the-art techniques relying on photometric stereo and/or deep learning to estimate surface normals from images. We first compare to the calibrated PS technique from sec.~\ref{sec:lambertian-ps}. While it is an improvement over the method of Yu et al.~\cite{yu-iccp-13}, we still refer to it as \cite{yu-iccp-13} in figs.~\ref{fig:results-qualitative} and \ref{fig:results-quantitative}. 
We also compare to the semi-calibrated method of Jung et al.~\cite{jung-ijcv-19}, which requires only knowledge of the camera geolocation. For deep learning techniques, we compare to the recent Deep Photometric Stereo Network (DPSN)~\cite{santo-iccv-17}, which operates on one pixel at a time. Since it assumes known point light source lighting, we re-trained this model using the sun position from a geographical location and date representative of our training dataset. In addition, we also compare to single image networks: Eigen and Fergus~\cite{eigen-iccv-15} and MarrNet~\cite{wu-nips-17}. Since they rely on a single image, we take the mean of their results averaged over all 8 inputs. 

The comparative results, shown qualitatively in fig.~\ref{fig:results-qualitative} and quantitatively in fig.~\ref{fig:results-quantitative}, clearly demonstrate that our approach significantly outperforms all other techniques. We observe that both single image techniques do not work well and result in very high median errors of around $40\degr$ and $70\degr$ for~\cite{wu-nips-17} and \cite{eigen-iccv-15}, respectively. For \cite{eigen-iccv-15}, this is probably due to the fact that they cannot handle the harsh shadows created by outdoor lighting during sunny days, since they train with indoor lighting only. In addition, MarrNet~\cite{wu-nips-17} outputs a voxel occupation grid and only produces normals as a byproduct (in its latent stage). As such, this method may not be fully optimized for normal estimation.

The PS techniques yield much better results but still yield quite significant error since sunny days do not contain sufficient constraints to accurately recover surface normals. The (improved) calibrated method of Yu et al.~\cite{yu-iccp-13} is comparable to the results obtained by DPSN, with a median normal angular estimation error of $33\degr$. Interestingly, the method of Jung et al.~\cite{jung-ijcv-19} actually yields better results with a median error of $22\degr$, despite needing less information (geolocation and time) than the calibrated methods. This could be due to its reliance on a parametric clear sky model to estimate lighting, which closely matches the actual ground truth lighting, and to its reliance on an intensity profile matching algorithm.

Note that most PS techniques capture with some degree of success the left/right component of the surface normals (roughly speaking, the red and blue tints in the normal maps). This axis is the same as the sun trajectory through the day when the camera is facing North or South. This results in strong photometric constraints on this axis. On the other hand, the recovery of the up/down axis is much less successful on most techniques as outdoor photometric cues lack information in this direction through a single sunny day. 

In contrast, our method yields a normal map that is, although a bit smoother, qualitatively very similar to the ground truth. Quantitatively, our approach achieves a median error of $14\degr$ over the evaluation set, with error predominantly below that of the second best performing method~\cite{jung-ijcv-19} (see fig.~\ref{fig:results-quantitative}). Even when trained on purely synthetic data, our network is able to generalize well to images rendered with real lighting. The difference in performance with respect to DPSN shows the usefulness of dealing with image patches, which allows the network to learn appropriate patch-based shape priors which can be exploited when the photometric cue alone is not sufficient. 

\subsection{Results on real captures}

We further evaluate our method on real data. We captured sequences of 8 outdoor images of 4 real statuettes during a single sunny day using a tripod-mounted Canon EOS 5D Mark III camera with a 300mm lens. These HDR images were obtained by merging camera exposure range from 1/8000 to 1 second at f/45. Ground-truth normals were obtained from a Creaform GO!SCAN 3D laser scans of the real objects. The results shown in fig.~\ref{fig:results-real} demonstrate the performance of our proposed method on such ill-posed outdoor PS problem. Using photometric cues alone, the two top statuettes from fig.~\ref{fig:results-real} have a maximum median reconstruction error of 29$\degr$ (owl) and 47$\degr$ (bust) due to the lighting matrix being nearly singular. In addition to relaxing the calibration requirements (as full environment maps are not needed for our technique), our learning-based technique improves the median surface reconstruction accuracy by up to 68\%.

\newcommand{\realresheight}{1.76cm}
\begin{figure}[t]
\centering
\begin{tabular}{@{}c@{}c@{}c@{}r@{}}
\includegraphics[height=\realresheight]{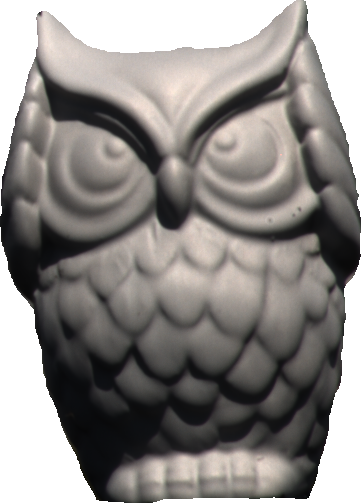} &
\includegraphics[height=\realresheight]{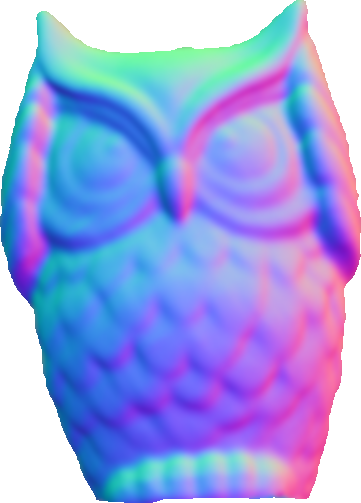} &
\includegraphics[height=\realresheight]{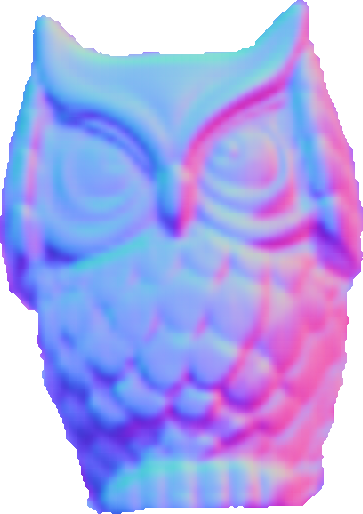} &
\makecell[tc]{\includegraphics[height=\realresheight]{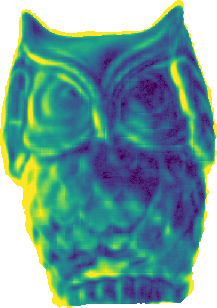}
\includegraphics[height=\realresheight]{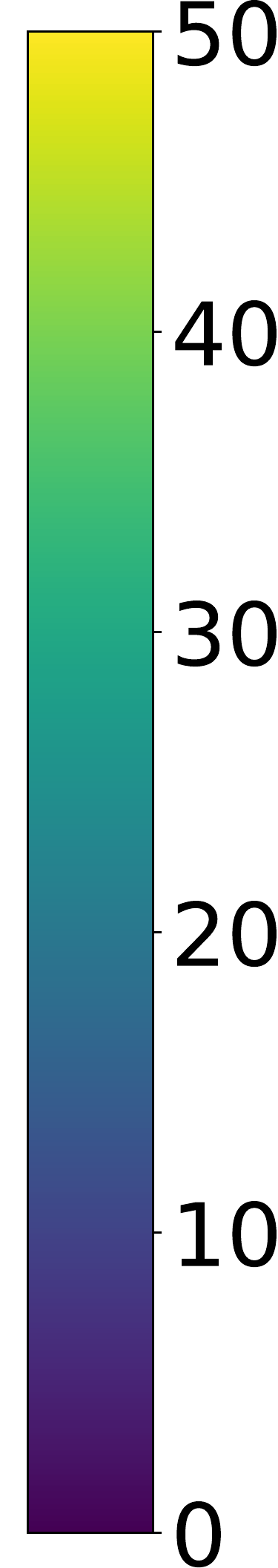} \vspace{-4pt}\\
\scriptsize 19.6$\degr$ / 79 } \\
\includegraphics[height=\realresheight]{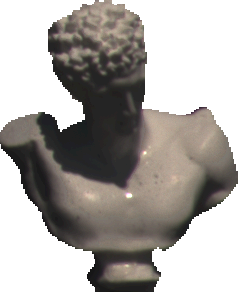} &
\includegraphics[height=\realresheight]{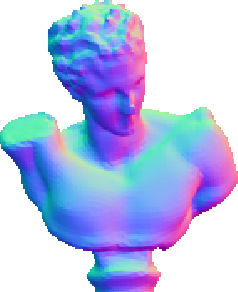} &
\includegraphics[height=\realresheight]{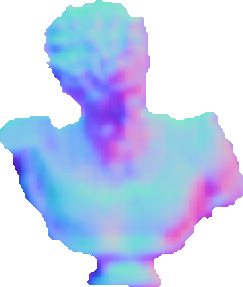} &
\makecell[tc]{\includegraphics[height=\realresheight]{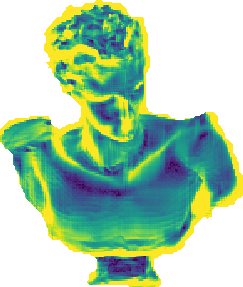}
\includegraphics[height=\realresheight]{figures/deep-realdata-20150822/colorbar_v.pdf} \vspace{-4pt}\\
\scriptsize 27.9$\degr$\ / 55.3 } \\
\includegraphics[height=\realresheight]{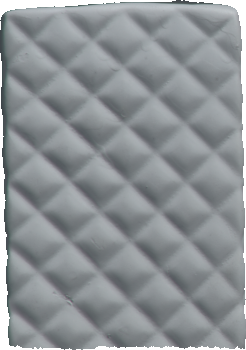} &
\includegraphics[height=\realresheight]{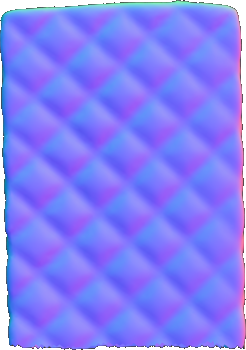} &
\includegraphics[height=\realresheight]{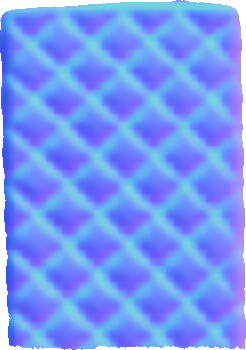} &
\makecell[tc]{\includegraphics[height=\realresheight]{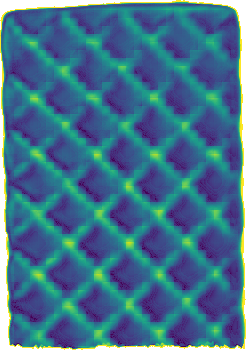}
\includegraphics[height=\realresheight]{figures/deep-realdata-20150822/colorbar_v.pdf} \vspace{-4pt}\\
\scriptsize 14.9$\degr$ / 91.9 } \\
\includegraphics[height=\realresheight]{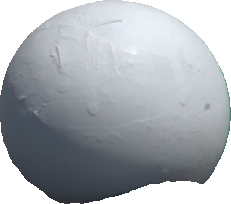} &
\includegraphics[height=\realresheight]{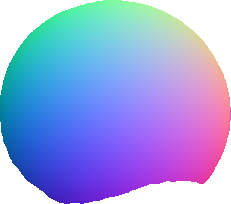} &
\includegraphics[height=\realresheight]{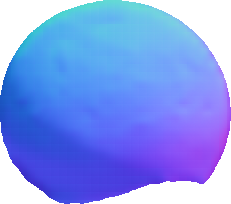} &
\makecell[tc]{\includegraphics[height=\realresheight]{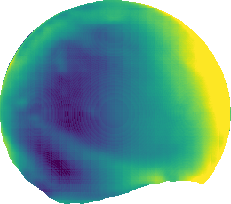}
\includegraphics[height=\realresheight]{figures/deep-realdata-20150822/colorbar_v.pdf} \vspace{-4pt}\\
\scriptsize 23.4$\degr$ / 68.4 } \\
\includegraphics[height=\realresheight]{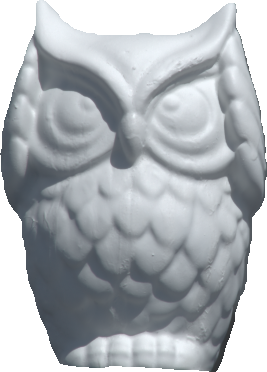} &
\includegraphics[height=\realresheight]{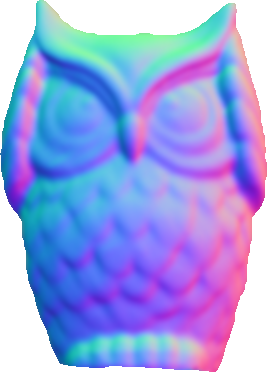} &
\includegraphics[height=\realresheight]{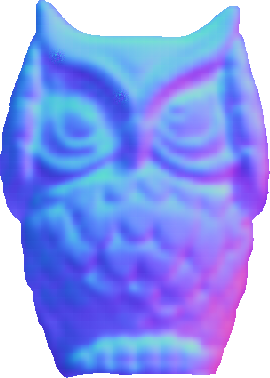} &
\makecell[tc]{\includegraphics[height=\realresheight]{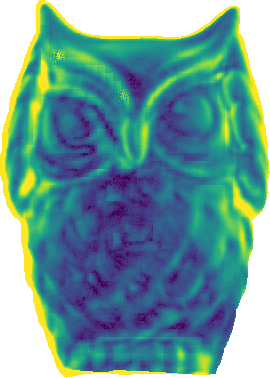}
\includegraphics[height=\realresheight]{figures/deep-realdata-20150822/colorbar_v.pdf} \vspace{-4pt}\\
\scriptsize 18$\degr$ / 80.1 } \\
(a) & (b) & (c) & \multicolumn{1}{c}{(d)} \\
\end{tabular}
\caption{Result on real statuettes (ill-posed, single day PS problem): (a)~example input images around solar noon; (b) the ground-truth (3D-scanned) normals; (c) normals estimated by our method; and (d) angular error map, median error in degree and amount of estimated normals within 30$\degr$~of the ground truth (R30). The top two rows were taken on 2015-08-22 while the bottom three rows were taken on 2018-05-24.}
\label{fig:results-real}
\end{figure}

\subsection{Analysis}
\label{sec:analysis}

We now analyze further our network, and in particular explore the robustness of our network to departures from the assumptions that were made in sec.~\ref{sec:proposed_method}. \textbf{More analysis is available in the supplementary material, including results on a partially cloudy day.}

\begin{figure*}[!t]
\centering
\includegraphics[width=1.0\linewidth]{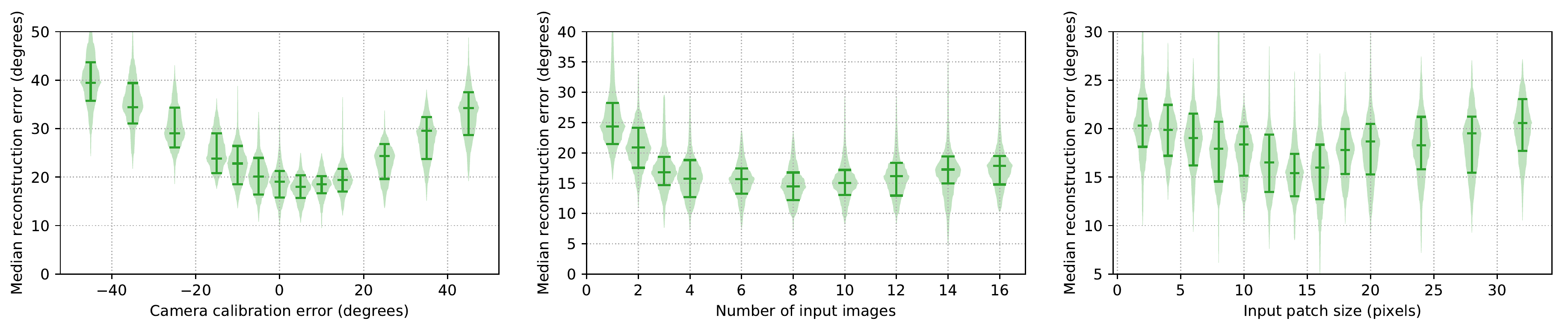}
\vspace{-1em}
\caption{(left) Median normal estimation error as box-percentile plots (see fig.~\ref{fig:results-quantitative}) in function of the camera deviation from north in degrees on our real lighting evaluation set. Positive means camera looking westward, negative means camera looking eastward. 
(center) Normal estimation error as box-percentile plots on our evaluation dataset in function of the number of input images $T$.
(right) Ablative study on the number of pixels in input.}
\label{fig:ablative_figures}
\end{figure*}

\subsubsection{Camera calibration error}

The impact on reconstruction performance when the north-facing camera hypothesis is infringed was studied by rotating the real environment maps used to render the evaluation dataset (sec.~\ref{sec:evaluation_dataset}), and show the results of this experiment in fig.~\ref{fig:ablative_figures}~(left). The slight improvement around $5\degr$ west calibration error is due to the timestamps of our real lighting dataset that are not perfectly aligned with the neural network expected timestamps. We observe that the median reconstruction error increases of approximately $5\degr$ per $10\degr$ error on camera calibration, showing that the network has some built-in robustness to these errors.


\subsubsection{Number of input images}
\label{sec:ablation_study}

We also investigated normal estimation performance in function of the number of inputs $T$ to the CNN (see sec.~\ref{sec:architecture}). Results ranging from a single input image ($T=1$, effectively performing shape-from-shading) to $T=16$ input images all uniformly taken from 9:00 to 16:00 are shown in fig.~\ref{fig:ablative_figures}~(center). We observe an rapid improvement in performance from one to three images, which is coherent with PS theory~\cite{woodham-opteng-80}. Performance continues to increase until $T=8$, probably because added constraints improves robustness to noise and non-diffuse materials. With $T > 8$, normal estimation error increases slightly. This could be due to an increase in the number of parameters to train in our model (the output tensor after concatenation is of dimension $14 \times 14 \times 32T$, thereby increasing the number of parameters in the second convolutional layer), making the model harder to train.

\subsubsection{Patch size}
\label{sec:patch-size}

Fig.~\ref{fig:ablative_figures}~(right) shows the impact of varying the patch size $P$. To achieve this, we add an adaptive max pooling layer of size $4 \times 4$ after the last residual block (see fig.~\ref{fig:architecture}). Accuracy increases with patch size until roughly 14 pixels, and then decreases. We hypothesize that very small patch sizes do not contain enough spatial context while too large patches reveal macroscopic object features, which the neural network fails to recognize in the new shapes of the test set.

\subsection{Limitations}

\begin{figure}
\begin{tabular}{c@{\extracolsep{\fill}}c@{}c@{}c@{}}
\includegraphics[width=0.31\linewidth]{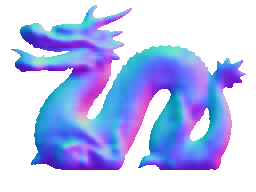} &
\includegraphics[width=0.31\linewidth]{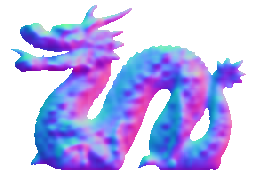} &
\includegraphics[width=0.31\linewidth]{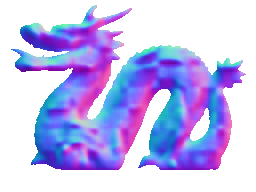} \\
\includegraphics[width=0.31\linewidth]{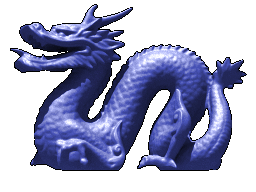} &
\includegraphics[width=0.31\linewidth]{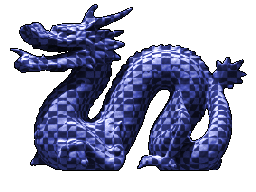} &
\includegraphics[width=0.31\linewidth]{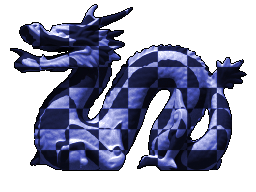} 
\end{tabular}
\caption{Limitation of our approach. Our network is trained on spatially uniform BRDFs, so testing it on spatially-varying albedo maps increases the estimation error. (left) Spatially-uniform albedos results in low error, while checkerboard albedo maps with (center) small and (right) large patterns increase the error.}\label{fig:limitations}
\end{figure}

The first limitation of the proposed method is that the camera is assumed to be pointing north. Although the network shows some resilience to errors in camera calibration (see fig.~\ref{fig:ablative_figures}), larger deviations from the assumed direction yield degraded performance. One possible way to circumvent this limitation would be to train direction-specific models and select the right one by detecting the camera orientation. Furthermore, while our approach is robust to non-Lambertian reflections, it assumes the scene to have a spatially-uniform BRDF. This assumption is shared with recent techniques like~\cite{mo-cvpr-2018}. Fig.~\ref{fig:limitations} shows the behavior with a spatially-varying BRDF composed of a checkerboard pattern with small and large squares. Unsurprisingly, the resulting normal maps appear distorted since the constant albedo assumption is broken. One interesting direction for future work here would be to train a network on the \emph{ratio} between pairs of images (e.g. as in~\cite{yu-iccp-13}), which effectively cancels out the albedo.

\section{Discussion}

This paper has presented a thorough analysis of outdoor PS under various illumination conditions captured over the course of a single day. In this scenario, we have no control over illumination, so existing methods for setting up optimal lighting~\cite{drbohlav-iccv-05,klaudiny-prl-14} cannot be applied. Through a data-driven analysis of the expected behavior of outdoor PS, we reveal natural factors that distinguish good and unfavorable daylight conditions and identify mainly two different types of working weather conditions: partially cloudy and clear days. Our analysis shows that occlusion of the sun by clouds provides additional photometric cues that improve the accuracy of the surface reconstruction. Furthermore, this improvement in conditioning can be observed in short time intervals and varies in accord with surface orientation.

However, with a cloudless sky, outdoor PS becomes ill-conditioned (even in case of simple Lambertian reflectance) and cannot be solved from photometric cues alone. To address this issue, we augment the available photometric cues with learned priors. As such, we present the first method for single-day outdoor PS based on deep learning. This new method is not limited to Lambertian objects and is also robust to shadows and specular highlights. It significantly outperforms previous work on a challenging evaluation dataset of virtual objects (lit by real sunny lighting conditions) and yields successful surface reconstructions on real objects.

One exciting direction for future work is to leverage the findings in this paper to design a unified approach for outdoor PS under skies with any amount of cloud coverage. For this, a properly-trained neural network could learn to reconstruct the detailed surface of a large class of objects observed under variable (but uncontrolled) natural outdoor illumination. Designing such an approach will however require care as simply training on clear days cannot reach the same performance on other weather conditions (see the supplementary material). We hypothesize that a hybrid approach, combining the photometric cues from sec.~\ref{sec:LambertianCalibrated} with a deep CNN such as the one in sec.~\ref{sec:NonLambertianSemiCalibrated} could be successful. A question that remains open to investigation is the adequate requirement in terms of lighting calibration as to provide beneficial information during reconstruction while also allowing for application in the wild. We believe the analysis presented in this paper sets the stage for exciting future work.

\bibliographystyle{IEEEtran}
\bibliography{main}

\vspace{-4em}
\begin{IEEEbiography}[{\includegraphics[width=1in,height=1.25in,clip,keepaspectratio]{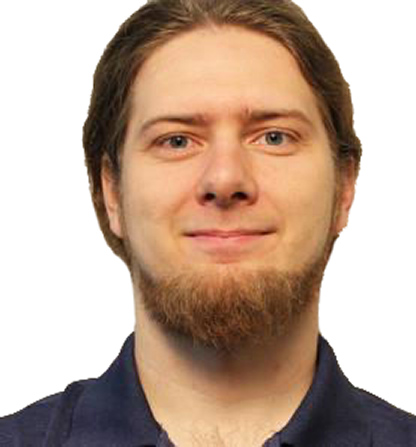}}]{Yannick Hold-Geoffroy} is a Research Engineer with Adobe in San Jos\'{e}. He received his Ph.D. degree in Electrical Engineering with a mention on the dean's honor roll from Universit\'{e} Laval, Canada, in 2018. He was awarded the CIPPRS Doctoral Dissertation Award in 2019. His research interests are in the understanding of natural images through machine learning and lighting modelling and estimation. 
\vspace{-5em}
\end{IEEEbiography}

\begin{IEEEbiography}[{\includegraphics[width=1in,height=1.25in,clip,keepaspectratio]{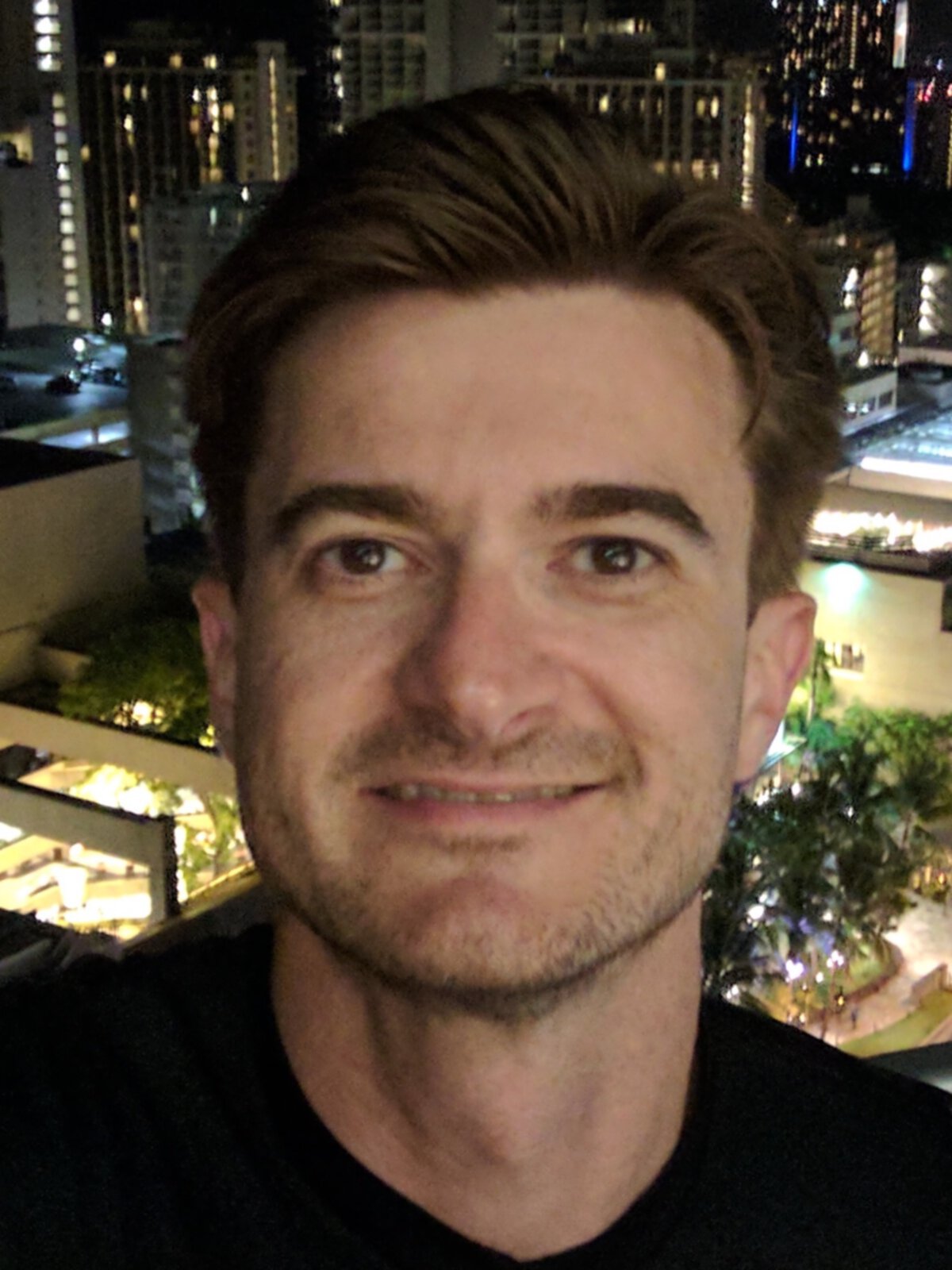}}]{Paulo Gotardo} is a Senior Research Scientist with Disney Research Studios in Zurich. His research on computer vision, graphics and machine learning focuses on modeling and capturing 3D geometry, motion, appearance and illumination in dynamic scenes, with application in building digital humans (movies, games, AR and VR). He received his B.Sc. and M.Sc. degrees in Informatics from the Federal University of Parana, Brazil, and his Ph.D. degree in Electrical and Computer Engineering from The Ohio State University. Before joining Disney Research, he was a research associate with the Advanced Computing Center for the Arts and Design and a Post-Doctoral Researcher with the Computational Biology and Cognitive Science Lab, both at Ohio State. Paulo was also an Associate Research Scientist with Disney Research Pittsburgh on the Carnegie Mellon University campus.

\vspace{-5em}
\end{IEEEbiography}

\begin{IEEEbiography}[{\includegraphics[width=1in,height=1.25in,clip,keepaspectratio]{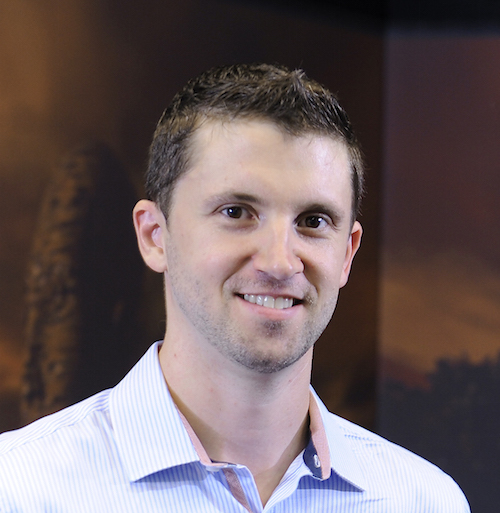}}]{Jean-Fran\c{c}ois Lalonde} is an Associate Professor in the ECE department at Universit\'{e} Laval, Canada. Previously, he was a Post-Doctoral Associate at Disney Research, Pittsburgh. He received his M.S. and Ph.D. from Carnegie Mellon in 2006 and 2011 respectively. His Ph.D. thesis won the 2010-11 CMU School of Computer Science Distinguished Dissertation Award. His research interests are in computer vision and deep learning, with a particular focus on lighting estimation, 3D reconstruction, tracking, and augmented reality. 

\end{IEEEbiography}





\end{document}